# AI Methods in Algorithmic Composition:
# A Comprehensive Survey

**Jose David Fernández**                                    JOSEDAVID@GEB.UMA.ES
**Francisco Vico**                                              FJV@GEB.UMA.ES
*Universidad de Málaga, Calle Severo Ochoa, 4, 119*
*Campanillas, Málaga, 29590 Spain*

## Abstract

Algorithmic composition is the partial or total automation of the process of music composition by using computers. Since the 1950s, different computational techniques related to Artificial Intelligence have been used for algorithmic composition, including grammatical representations, probabilistic methods, neural networks, symbolic rule-based systems, constraint programming and evolutionary algorithms. This survey aims to be a comprehensive account of research on algorithmic composition, presenting a thorough view of the field for researchers in Artificial Intelligence.

## 1. Introduction

Many overly optimistic, but ultimately unfulfilled predictions were made in the early days of Artificial Intelligence, when computers able to pass the Turing test seemed a few decades away. However, the field of Artificial Intelligence has grown and got matured, developing from academic research and reaching many industrial applications. At the same time, key projects and challenges have captivated public attention, such as driverless cars, natural language and speech processing, and computer players for board games.

The introduction of formal methods have been instrumental in the consolidation of many areas of Artificial Intelligence. However, this presents a disadvantage for areas whose subject matter is difficult to define in formal terms, which naturally tend to become marginalized. That is the case of *Computational Creativity* (also known as *Artificial Creativity*), which can be loosely defined as the computational analysis and/or synthesis of works of art, in a partially or fully automated way. Compounding the problem of marginalization, the two communities naturally interested in this field (AI and the arts) speak different languages (sometimes *very* different!) and have different methods and goals[1], creating great difficulties in the collaboration and exchange of ideas between them. In spite of this, small and sometimes fragmented communities are active in the research of different aspects of Computational Creativity.

The purpose of this survey is to review and bring together existing research on a specific style of Computational Creativity: *algorithmic composition.* Interpreted literally, algorithmic composition is a self-explanatory term: the use of algorithms to compose music. This is a very broad definition, because for centuries musicians have been proposing methods that can be considered as algorithmic in some sense, even if human creativity plays a key role. Some commonly cited examples include d'Arezzo's *Micrologus*, species counterpoint, Mozart's dice games, Schoenberg's twelve-tone technique, or Cage's aleatoric music. Readers interested in these and other pre-computer examples of algorithmic composition are referred to the introductory chapters of almost any thesis or book on the subject, such as Díaz-Jerez's (2000), Aschauer's (2008) or Nierhaus's (2009). In this survey, we will use the term algorithmic composition in a more restricted way, as the partial or total automation of music composition by formal, computational means. Of course, pre-computer examples of algorithmic composition can be implemented on a computer, and some of the approaches reviewed in this survey implement a classical methodology. In general, the focus will be on AI techniques, but self-similarity and cellular automata will also be reviewed as modern computational techniques that can be used for generating music material without creative human input.

---

1. Related to this problem, it is not uncommon for engineering concepts to become bent in strange ways when interpreted by artists. See Footnote 28 in page 550 for a particularly remarkable example.

## 1.1 Motivation

Some useful starting points for researching the past and present of computer music are the *Computer Music Journal*, the *International Computer Music Conference*[2] annually organized by the *International Computer Music Association*[3], and some books such as *Machine Models of Music* (Schwanauer & Levitt, 1993), *Understanding music with AI* (Balaban et al., 1992), *Music and Connectionism* (Todd & Loy, 1991), and the anthologies of selected articles from the *Computer Music Journal* (Roads & Strawn, 1985; Roads, 1992). However, these resources are not only about algorithmic composition, but computer music in general. For more specific information on algorithmic composition, surveys are a better option.

There are many surveys reviewing work on algorithmic composition. Some review both analysis and composition by computer with AI methods (Roads, 1985), while others discuss algorithmic composition from a point of view related to music theory and artistic considerations (Collins, 2009), or from the personal perspective of a composer (Langston, 1989; Dobrian, 1993; Pope, 1995; Maurer, 1999). Some of them provide an in depth and comprehensive view of a specific technique for algorithmic composition, as Anders and Miranda (2011) do for constraint programming, as Ames (1989) does for Markov chains, or as Santos et al. (2000) do for evolutionary techniques, while some others are specialized in the comparison between paradigms for computational research on music, as Toiviainen (2000). Others offer a wide-angle (but relatively shallow) panoramic of the field (Papadopoulos & Wiggins, 1999), review the early history of the field (Loy & Abbott, 1985; Ames, 1987; Burns, 1994), or analyze methodologies and motivations for algorithmic composition (Pearce et al., 2002). There are also works that combine in depth and comprehensive reviews for a wide range of methods for algorithmic composition, such as Nierhaus's (2009) book.

In this context, a natural question arises: why yet another survey? The answer is that no existing survey article fulfills the following criteria: (a) to cover all methods in a comprehensive way, but from a point of view primarily focused on AI research, and (b) to be centered on algorithmic composition.[4] Nierhaus's (2009) book on algorithmic

---

composition comes close to fulfilling these criteria with long, detailed expositions for each method and comprehensive reviews of the state of the art. In contrast, this survey is intended to be a reasonably short article, without lengthy descriptions: just a reference guide for AI researchers. With these aims in mind, this survey is primarily structured around the methods used to implement algorithmic composition systems, though early systems will also be reviewed separately.

A second, more practical motivation is accessibility. Since Computational Creativity balances on the edge between AI and the arts, the relevant literature is scattered across many different journals and scholarly books, with a broad spectrum of topics from computer science to music theory. As a very unfortunate consequence, there are *many* different paywalls between researchers and relevant content, translating sometimes into a lot of hassle, only partially mitigated by relatively recent trends like self-archiving. This survey brings together a substantial body of research on algorithmic composition, with the intention of conveying it more effectively to AI researchers.

## 2. Introducing Algorithmic Composition

Traditionally, composing music has involved a series of activities, such as the definition of melody and rhythm, harmonization, writing counterpoint or voice-leading, arrangement or orchestration, and engraving (notation). Obviously, this list is not intended to be exhaustive or readily applicable to every form of music, but it is a reasonable starting point, especially for classical music. All of these activities can be automated by computer to varying degrees, and some techniques or languages are more suitable for some of these than others (Loy & Abbott, 1985; Pope, 1993).

For relatively small degrees of automation, the focus is on languages, frameworks and graphical tools to provide support for very specific and/or monotone tasks in the composition process, or to provide raw material for composers, in order to bootstrap the composition process, as a source of inspiration. This is commonly known as *computer-aided algorithmic composition* (CAAC), and constitutes a very active area of research and commercial software development: many software packages and programming environments can be adapted to this purpose, such as SuperCollider (McCartney, 2002), Csound (Boulanger, 2000), MAX/MSP (Puckette, 2002), Kyma (Scaletti, 2002), Nyquist (Simoni & Dannenberg, 2013) or the AC Toolbox (Berg, 2011). The development of experimental CAAC systems at the IRCAM[5] (such as PatchWork, OpenMusic and their various extensions) should also be emphasized (Assayag et al., 1999). Ariza's comprehensive repository of software tools and research resources for algorithmic composition[6] constitutes a good starting point (Ariza, 2005a) to explore this ecosystem, as well as algorithmic composition in general. Earlier surveys (such as Pennycook, 1985 and Pope, 1986) are also useful for understanding the evolution of the field, especially the evolution of graphical tools to aid composers.

Our survey, on the other hand, is more concerned with algorithmic composition with higher degrees of automation of compositional activities, rather than typical CAAC. In other words, we focus more on techniques, languages or tools to computationally encode human musical creativity or automatically carry out creative compositional tasks with minimal or no human intervention, instead of languages or tools whose primary aim is to aid human composers in their own creative processes.

---

5. http://www.ircam.fr/
6. http://www.flexatone.net/algoNet/

Obviously, the divide between both ends of the spectrum of automation (CAAC representing a low degree of automation, algorithmic composition a high degree of automation) is not clear, because any method that automates the generation of creative works can be used as a tool to aid composers, and systems with higher degrees of automation can be custom-built on top of many CAAC frameworks.[7] Furthermore, a human composer can naturally include computer languages and tools as an integral part of the composition process, such as Brian Eno's concept of *generative music* (Eno, 1996). To conclude these considerations, this survey is about computer systems for automating compositional tasks where the user is not expected to be the main source of creativity (at most, the user is expected to set parameters for the creative process, encode knowledge about how to compose, or to provide examples of music composed by humans to be processed by the computer). This also includes real-time automatic systems for music improvisation, such as in jazz performance, or experimental musical instruments that automate to a certain extent the improvisation of music.

Finally, a few more considerations, to describe what this survey is not about:

- Although music can be defined as "organized sound", a composition written in traditional staff notation does not fully specify how the music actually sounds: when a piece of music is performed, musicians add patterns of small deviations and nuances in pitch, timing and other musical parameters. These patterns account for the musical concept of *expressiveness* or *gesture*, and they are necessary for the music to sound natural. While the problem of automatically generating expressive music is important in itself, and involves creativity, it is clearly not within the boundaries of algorithmic composition as reviewed in this survey. The reader is referred to Kirke and Miranda's (2009) review of this area for further information.

- The computational synthesis of musical sounds, or algorithmic sound synthesis, can be understood as the logical extension of algorithmic composition to small timescales; it involves the use of languages or tools for specifying and synthesizing sound waveforms, rather than the more abstract specification of music associated with traditional staff notation. The line between algorithmic composition and algorithmic sound synthesis is blurred in most of the previously mentioned CAAC systems, but this survey is not concerned with sound synthesis; interested readers may refer to Roads's (2004) book on the subject.

- In computer games (and other interactive settings), music is frequently required to gracefully adapt to the state of the game, according to some rules. This kind of music is commonly referred to as *non-linear music* (Buttram, 2003) or *procedural audio* (Farnell, 2007). Composing non-linear music presents challenges of its own, not specifically related to the problem of algorithmic composition, so we will not review the literature on this kind of music.

---

7. This is the case of many of the systems for algorithmic composition described here. For example, PWConstraints (described in Section 3.2.3) is built on top of PatchWork, as described by Assayag et al. (1999).

These three scenarios (automated expressiveness, algorithmic sound synthesis and non-linear music) will be sparingly mentioned in this survey, only mentioned when innovative (or otherwise notable) techniques are involved.

## 2.1 The Early Years

In this section, we will review early research published on algorithmic composition with computers, or with a clear computational approach. While these references might have been discussed by methodology in the following sections, it is useful to group them together here, since it is difficult to find a survey discussing all of them.

The earliest use of computers to compose music dates back to the mid-1950s, roughly at the same time as the concept of Artificial Intelligence was coined at the Darmouth Conference, though the two fields did not converge until some time later. Computers were expensive and slow, and also difficult to use, as they were operated in batch mode.

One of the most commonly cited examples is Hiller and Isaacson's (1958) *Illiac Suite*, a composition that was generated using rule systems and Markov chains, late in 1956. It was designed as a series of experiments on formal music composition. During the following decade, Hiller's work inspired colleagues from the same university to further experiment with algorithmic composition, using a library of computer subroutines for algorithmic composition written by Baker (also a collaborator of Hiller), MUSICOMP (Ames, 1987). This library provided a standard implementation of the various methods used by Hiller and others.

Iannis Xenakis, a renowned *avant-garde* composer, profusely used stochastic algorithms to generate raw material for his compositions, using computers since the early 1960s to automate these methods (Ames, 1987). Though his work can be better described as CAAC, he still deserves being mentioned for being a pioneer. Koenig, while not as well known as Xenakis, also was a composer that in 1964 implemented an algorithm (PROJECT1) using serial composition (a musical theory) and other techniques (as Markov chains) to automate the generation of music (Ames, 1987).

However, there were also several other early examples of algorithmic composition, though not so profusely cited as Hiller and Xenakis's. *Push Button Bertha*, composed in 1956 (Ames, 1987) around the same time as Hiller's *Illiac Suite*, is perhaps the third most cited example: a song whose music was algorithmically composed as a publicity stunt by Burroughs (an early computer company), generating music similar to a previously analyzed corpus. However, there is at least one earlier, unpublished work by Caplin and Prinz in 1955 (Ariza, 2011), which used two approaches: an implementation of Mozart's dice dame and a generator of melodic lines using stochastic transitional probabilities for various aspects of the composition. Another commonly cited example by Brooks et al. (1957) explored the potential of the *Markoff*[8] *chain* method.

Several other early examples are also notable. Olson's (1961) dedicated computer was able to compose new melodies related to previously fed ones, using Markov processes. While the work was submitted for publication in 1960, they claimed to have built the machine in the early 1950s. Also of interest is Gill's (1963) algorithm, implemented at the request of the BBC, which represents a hallmark in the application of classical AI techniques to algorithmic composition: it used a hierarchical search with backtracking to guide a compositional process inspired by Schoenberg's twelve-tone technique. Finally, it is worth mentioning what may represent the first dissertation on algorithmic composition: Padberg's (1964) Ph.D. thesis implemented a compositional framework (based in formal music theory) in computer code. Her work is unusual in that, instead of using random number generators, she used raw text input to drive procedural techniques in order to generate all the parameters of the composition system.

---

8. "Markov" and "Markoff" are alternative transliterations of the Russian surname Ма́рков. The spelling "Markov" has been prevalent for decades, but many older papers used "Markoff".

Non-scholarly early examples also exist, though they are difficult to assess because of the sparsity of the published material, and the fact that they are mostly not peer-reviewed. For example, Pinkerton (1956) described in *Scientific American* a "Banal Tune-Maker", a simple Markov chain created from several tens of nursery tunes, while Sowa (1956) used a GENIAC machine[9] to implement the same idea (Cohen, 1962), and Raymond Kurzweil implemented in 1965 (Rennie, 2010) a custom-made device that generated music in the style of classical composers. Another example, unfortunately shrouded in mystery, is Raymond Scott's "Electronium" (Chusid, 1999), an electronic device whose development spanned several decades, reportedly able to generate abstract compositions. Unfortunately, Scott never published or otherwise explained his work.

As machines became less expensive, more powerful and in some cases interactive, algorithmic composition slowly took off. However, aside from the researchers at Urbana (Hiller's university), there was little continuity in research, and reinventing the wheel in algorithmic composition techniques was common. This problem was compounded by the fact that initiatives in algorithmic composition often came from artists, who tended to develop *ad hoc* solutions, and the communication with computer scientists was difficult in many cases.

## 3. The Methods

The range of methodological approaches used to implement algorithmic composition is notably wide, encompassing many, very different methods from Artificial Intelligence, but also borrowing mathematical models from Complex Systems and even Artificial Life. This survey has been structured by methodology, devoting a subsection to each one:

> 3.1 **Grammars**
>
> 3.2 **Symbolic, Knowledge-Based Systems**
>
> 3.3 **Markov Chains**
>
> 3.4 **Artificial Neural Networks**
>
> 3.5 **Evolutionary and Other Population-Based Methods**
>
> 3.6 **Self-Similarity and Cellular Automata**

Figure 1 summarizes the taxonomy of the methods reviewed in this survey. Together, Sections 3.1 and 3.2 describe work using symbolic techniques that can be characterized as classical "good old-fashioned AI". Although grammars (Section 3.1) are symbolic and knowledge-based, and thus should be included as part of Section 3.2, they have been segregated in a separate subsection because of their relative historical importance in algorithmic composition. Sections 3.3 and 3.4 describe work using various methodologies for machine learning, and Section 3.5 does the same for evolutionary algorithms and other population-based optimization methods. Although the methodologies described in Section 3.6 are not really a form of "Artificial Intelligence", they have been included because of their importance in algorithmic composition as automatic sources of music material (i.e., they do not depend on any model of human creativity for generating music material).

---

9. A GENIAC *Electric Brain*, an electric-mechanic machine promoted as an educational toy. Despite being marketed as a computer device, all the "computing" was performed by the human operator.

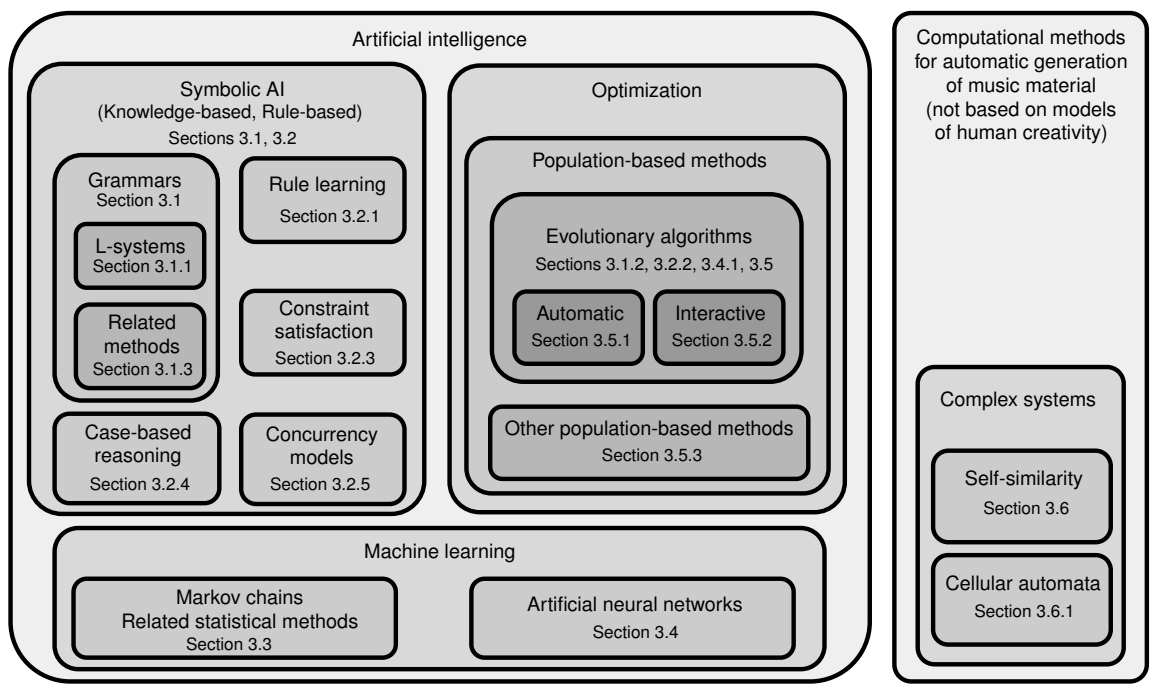

Figure 1: Taxonomy of the methods reviewed in this survey

There have been other attempts to systematize algorithmic composition, such as the taxonomies of Papadopoulos and Wiggins (1999) and Nierhaus (2009). Our taxonomy is roughly similar to Nierhaus's, with some differences, such as including L-systems as grammars instead of self-similar systems. The reader may be surprised to find that many methods for machine learning and optimization are missing from our taxonomy. There are several reasons for this. In some cases, some methods are subsumed by others. For example, in machine learning, many different methods have been formulated in the mathematical framework of artificial neural networks. In other cases, a method has been used only rarely, almost always together with other methods. For example, in optimization, this is the case of *tabu search*, which has been used a few times in the context of constraint satisfaction problems (Section 3.2.3), and *simulated annealing*, which has been occasionally combined with constraint satisfaction, Markov processes and artificial neural networks.

It is difficult to neatly categorize the existing literature in algorithmic composition with any hierarchical taxonomy, because the methods are frequently hybridized, giving rise to many possible combinations. This is specially true for evolutionary methods, which have been combined with almost every other method. Additionally, some papers can be considered to belong to different methodologies, depending on the selected theoretical framework[10], while others are unique in their approaches[11], further complicating the issue. Finally, the lines between some methods (as rule systems, grammars and Markov chains) are frequently blurred: in some cases, ascribing a work to one of them becomes, in the end, a largely arbitrary exercise depending on the terminology, intentions and the domain of the researchers. Each method will be presented separately (but also presenting existing hybridizations with other methods), describing the state of the art in a mostly chronological order for each method.

Although this classification is not fully comprehensive, we have only found one (arguably remote) example using a method that is not related to the ones listed above: Amiot et al. (2006), who applied the Discrete Fourier Transform (DFT) to generate variations of musical rhythms. Given a rhythm as a sequence of numerical symbols, they represented it in the frequency domain by computing its DFT. Variations on that rhythm were generated by slightly perturbing the coefficients of the transform and converting back to the time domain.

## 3.1 Grammars and Related Methods

In broad terms, a formal grammar may be defined as a set of rules to expand high-level symbols into more detailed sequences of symbols (words) representing elements of formal languages. Words are generated by repeatedly applying rewriting rules, in a sequence of so-called derivation steps. In this way, grammars are suited to represent systems with hierarchical structure, which is reflected in the recursive application of the rules. As hierarchical structures can be recognized in most styles of music, it is hardly surprising that formal grammar theory has been applied to analyze and compose music for a long time[12], despite recurring concerns that grammars fail to capture the internal coherency and subtleties required for music composition (Moorer, 1972).

To compose music using formal grammars, an important step is to define the set of rules of the grammar, which will drive the generative process. The rules are traditionally multi-layered, defining several subsets (maybe even separated in distinct grammars) of rules for different phases of the composition process: from the general themes of the composition, down to the arrangement of individual notes. While early authors derived the rules by hand from principles grounded in music theory, other methods are possible, like examining a corpus of pre-existing musical compositions to distill a grammar able to generate compositions in the general style of the corpus, or using evolutionary algorithms. Another important aspect is the mapping between the formal grammar and the musical objects that it generates, which usually relates the symbols of the derived sequences with elements of the music composition, as notes, chords or melodic lines. However, other mappings are possible, as using the derivation tree to define the different aspects of the musical composition. Another important aspect of the automatic composition process is the election of the grammatical rules to be applied. While many approaches are possible, the use of activation probabilities for the rules (stochastic grammars) is common. In the process of compiling information for this survey, it has been noted that almost all research has been done on regular and context-free grammars, as context-sensitive and more general grammars seem to be very difficult to implement effectively, except for very simple toy systems.

---

10. For example, Markov chains can be formulated as stochastic grammars; some self-similar systems can be characterized as L-system grammars; rule learning and case-based reasoning are also machine learning methods; etc.

11. For example, Kohonen's method (Kohonen et al., 1991), which is neither grammatical nor neural nor Markovian, but can be framed in either way, according to its creator.

12. See, e.g., the survey by Roads (1979).

| Reference | Composition task | Comments |
|---|---|---|
| Lidov & Gabura, 1973 | melody | early proposal |
| Rader, 1974 | melody | early proposal, very detailed grammar |
| Ulrich, 1977 | jazz chord identification | integrated in an *ad hoc* system (to produce jazz improvisations) |
| Baroni & Jacoboni, 1978 | grammar for Bach chorales | early proposal |
| Leach & Fitch, 1995 (XComposer) | structure, rhythm and melody | uses chaotic non-linear systems (self-similarity) |
| Hamanaka et al., 2008 | generate variations on two melodies (by altering the derivation tree) | inspired by Lerdahl et al.'s (1983) GTTM |
| Roads, 1977 | structure, rhythm and melody | grammar compiler |
| Holtzman, 1981 | structure, rhythm and melody | grammar compiler |
| Jones, 1980 | structure | space grammars (uses the derivation tree) |
| Bel, 1992 (Bol Processor) | improvisation of *tabla* rhythms | tool for field research |
| Kippen & Bel, 1989 | improvisation of *tabla* rhythms | grammatical inference |
| Cruz-Alcázar & Vidal-Ruiz, 1998 | melody | grammatical inference |
| Gillick et al., 2009 | jazz improvisation | grammatical inference. Implemented as an extension to Keller and Morrison's (2007) ImprovGenerator |
| Kitani & Koike, 2010 (ImprovGenerator) | real-time drum rhythm improvisation | online grammatical inference |
| Keller & Morrison, 2007 (Impro-Visor) | jazz improvisation | sophisticated GUI interface |
| Quick, 2010 | classical three-voice counterpoint | integrated in a Schenkerian framework |
| Chemillier, 2004 | jazz chord sequences | implemented in OpenMusic and MAX |

Table 1: References for Section 3.1 (algorithmic composition with grammars), in order of appearance.

Lidov and Gabura (1973) implemented an early example of a formal grammar to compose simple rhythms. Another early example was implemented by Rader (1974): he defined a grammar by hand from rather simple music concepts, enriching the rules of the grammar with activation probabilities. Other early examples used grammars driven by rules from music theories, either as a small part of a synthesis engine, as Ulrich's (1977) grammar for enumerating jazz chords, or by inferring the rules from classical works, as Baroni and Jacoboni's (1978) grammar to generate melodies. *A Generative Theory of Tonal Music* (Lerdahl et al., 1983), a book presenting a grammatical analysis of tonal music, is a relatively early theoretical work that can be said to have influenced the use of grammars for algorithmic composition, though it is not directly concerned with algorithmic composition, but with a grammatical approach to the analysis of music. This book has been widely popular, and has had a lasting impact on the field and high citation rates. Examples of later work inspired by this book include Pope's (1991) "T-R Trees", Leach and Fitch's (1995) "event trees", and Hamanaka et al.'s (2008) "melody morphing".

In the 1980s, some proposed approaches more in line with computer science, abstracting the process to generate the grammars instead of codifying them by hand, though at the cost of producing less interesting compositions. Roads (1977) proposed a framework to define, process and use grammars to compose music, while Holtzman (1981) described a language to define music grammars and automatically compose music from them. Meanwhile, Jones (1980) proposed the concept of space grammars, in conjunction with a novel mapping technique: instead of using the terminal symbols as the building blocks of the composition, he used the derivation tree of the terminal sequence to define the characteristics of the composition. This approach was unfortunately not developed far enough to yield significant results. In spite of these early efforts, most research on grammatical representations of music was focused on analysis rather than synthesis. Some instances, such as Steedman's (1984) influential grammar for the analysis of jazz chord progressions, were later adapted for synthesis (see below).

The problem with a grammatical approach to algorithmic composition is the difficulty to manually define a set of grammatical rules to produce good compositions. This problem can be solved by generating the rules of the grammar (and the way they are applied) automatically. For example, although Bel (1992) implemented the BOL processor to facilitate the creation by hand of more or less sophisticated music grammars[13], he also explored the automated inference of regular grammars (Kippen & Bel, 1989). Later, Cruz-Alcázar and Vidal-Ruiz (1998) implemented several methods of *grammatical inference*: analyze a corpus of pre-existing classical music compositions, represented with a suitable set of symbols, then inducing stochastic regular grammars (Markov chains) able to parse the compositions in the corpus, and finally applying these grammars to generate new compositions that are in a similar style to the compositions in the corpus. Gillick et al. (2009) used a similar approach (also Markovian) to synthesize jazz solos, but with a more elaborated synthesis phase. Kitani and Koike (2010) provide another example of grammatical inference, in this case used for real-time improvised accompaniment.

However, others still designed their grammars by hand, carefully choosing the mapping between terminal symbols and musical objects, as Keller and Morrison (2007) did for jazz improvisations. Another approach is to take a pre-existing music theory with a strong hierarchical methodology, as designing a grammar inspired in Schenkerian analysis (Quick, 2010), or using Lerdhal's grammatical analysis to derive new compositions from two previously existing ones by altering the derivation tree (Hamanaka et al., 2008), or even de-veloping a jazz on-the-fly improviser (Chemillier, 2004) by adapting Steedman's grammar, previously implemented for analysis purposes.

---

13. Initially to represent and analyze informal knowledge about Indian *tabla* drumming, but later also to represent other music styles.

| Reference | Composition task | Comments |
|---|---|---|
| Prusinkiewicz, 1986 | melody | mapping turtle graphics to music scores |
| Nelson, 1996 | melody | mapping turtle graphics to music scores |
| Mason & Saffle, 1994 | melody (counterpoint is suggested) | mapping turtle graphics to music scores |
| Soddell & Soddell, 2000 | aural representations of biological data | L-system modulates pitch intervals |
| Morgan, 2007 | composition for a large instrumental ensemble | *ad hoc* symbolic mapping |
| Langston, 1989 | melody | L-system is interpreted to arrange pre-specified fragments |
| Worth & Stepney, 2005 | melody | several mappings and L-system types |
| Manousakis, 2006 | melody (sound synthesis) | complex, multi-dimensional mapping. Implemented in MAX. |
| McCormack, 1996 | melody, polyphonies | contex-sensitive L-systems |
| DuBois, 2003 | real-time accompaniment | implemented in MAX |
| Wilson, 2009 | melody | mapping turtle graphics to music scores |
| McGuire, 2006 | arpeggiator | simple symbolic mapping |
| Watson, 2008 | base chord progression | L-systems are used the context of a larger, multi-stage system |
| Gogins, 2006 | voice leading | Musical theory (pitch spaces). Implemented in Csound |
| Bulley & Jones, 2011 | arpeggiator | part of a real-time art installation. Implemented in MAX |
| Pestana, 2012 | real-time accompaniment | implemented in MAX |

Table 2: References for Section 3.1.1, in order of appearance.

### 3.1.1 L-Systems

Lindenmayer Systems, commonly abbreviated to L-systems, are a specific variant of formal grammar, whose most distinctive feature is parallel rewriting, i.e., at each derivation step, not one but all possible rewriting rules are applied at once. They have been successfully applied in different scenarios, specially to model microbial, fungi and plant growth and shapes, because they are particularly well-suited to represent the hierarchical self-similarity characteristic of these organisms. This ability to represent self-similar structures, together with the fact that L-systems are easier to understand and apply than traditional formal grammars, have made L-systems fairly popular in algorithmic composition.

Arguably, the most visually stunning way to use L-systems has been the synthesis of 2D and 3D renderings of plants, using a mapping from sequences of symbols to graphics based on turtle graphics (Prusinkiewicz & Lindenmayer, 1990). It is only natural that the first application of L-systems to algorithmic composition used turtle graphics to render an image that was then interpreted into a musical score (Prusinkiewicz, 1986), mapping coordinates, angles and edge lengths into musical objects. This approach has been used by music composers, as Nelson's (1996) *Summer Song* and Mason and Saffle's (1994) idea of using different rotations and stretchings of the image to implement counterpoint. As a funny side note, Soddell and Soddell (2000) generated aural renditions of their biological L-system models, to explore new ways to understand them. Additionally, other composers used new approaches not dependent upon the graphical interpretation of the L-systems, such as Morgan's (2007) symbolic mapping. One popular is to pre-generate a collection of short fragments and/or other musical objects, and define an algorithm to interpret the final sequence of symbols as instructions that transform and arrange the fragments into a composition. This approach has been used by Langston (1989) and Kyburz (Supper, 2001), while Edwards (2011) used a more convoluted but ultimately similar mapping.

However, these two approaches (the graphics-to-music and the pre-generated sequences) only scratch the surface of the technical possibilities to generate music with L-systems; many other mappings are possible (Worth & Stepney, 2005). In some cases, these mappings can become exceedingly complex, such as the implementation of Manousakis (2006), whose L-systems drove a multidimensional automata whose trajectory was then interpreted as music. While most composers and researchers experimented with context-free L-systems, McCormack (1996, 2003a) used context-sensitive, parametric L-systems to increase the expressiveness of the compositions and enable the implementation of polyphony. He also used a rich and comprehensive mapping from the symbol sequence to the musical score, interpreting the symbols in the sequence as instructions to modulate the parameters of an automata driving a MIDI synthesizer, though the grammars were ultimately specified by hand. DuBois (2003) used a simpler but also rich approach, mapping the symbols to elemental musical objects (as notes or instruments) or simple transformations applied to them, using brackets to encode polyphony. He also used L-systems to drive real-time synthetic accompaniment, by extracting features from the audio signal of a performer (as the pitch and loudness of the notes), encoding them as symbols to be expanded by L-system rules, and using the resulting symbol sequences to drive MIDI synthesizers. In spite of these developments, new mappings based on the images rendered by the turtle method are still investigated (Wilson, 2009).

L-systems can also be used to implement tools to assist the compositional process by solving just a part of it, as generating more complex arpeggios than off-the-shelf arpeggiators (McGuire, 2006), or providing just the base chord progression of a composition (Watson, 2008), sometimes applying elements of music theory to implement the rules (Gogins, 2006). Another area of research is the implementation of real-time improvisers, either for limited parts of the composition process (Bulley & Jones, 2011), or for accompaniment (Pestana, 2012).

### 3.1.2 Grammars and Evolutionary Algorithms

Evolutionary methods have also been used together with grammars. In this case, a common approach is to evolve the grammatical rules, as in *GeNotator* (Thywissen, 1999), in which the genomes are grammars specified through a GUI and the fitness function is interactive (the user assigns the fitness of the grammars). A more exotic example by Khalifa et al.

| Reference | Composition task | Comments |
|---|---|---|
| Thywissen, 1999 (GeNotator) | structure | the grammar is the genotype in an interactive evolutionary algorithm |
| Khalifa et al., 2007 | melody | the grammar is part of the fitness function |
| Ortega et al., 2002 | melody | grammatical evolution |
| Reddin et al., 2009 | melody | grammatical evolution |
| Shao et al., 2010 (Jive) | melody | interactive grammatical evolution |
| Bryden, 2006 | melody | interactive evolutionary algorithm with L-systems |
| Fox, 2006 | melody | interactive evolutionary algorithm with L-systems |
| Peck, 2011 | melody | evolutionary algorithm with L-systems |
| Dalhoum et al., 2008 | melody | grammatical evolution with L-systems |

Table 3: References for Section 3.1.2, in order of appearance.

(2007) uses the grammar as part of the fitness function instead of the generation of the compositions.

Some evolutionary methods are specifically adapted to handle grammars. This is the case of *grammatical evolution*, a method in which the genomes are sequences of numbers or symbols controlling the application of rules of a pre-defined (and possibly stochastic) grammar. The most common approach is to represent the music as the output from the grammar, which can range from very general to specific for a given music style. Several instances of this method have been developed: from an early, bare-bones implementation (Ortega et al., 2002) to a more elaborated one using a simple fitness function based on general concepts from music theory (Reddin et al., 2009). However, there are other approaches, such as the system implemented by Shao et al. (2010), whose grammar is used to produce intermediate code, which is then used to generate the music.

As in the more general case of formal grammars, evolutionary algorithms have been used to create L-systems. However, most examples use an interactive fitness function (the fitness is assigned by a human), like the basic implementation of Bryden (2006) and the approach based on *genetic programming* used by Fox (2006). Others use very simplistic fitness functions, with modest results (Peck, 2011). A more sophisticated approach was used by Dalhoum et al. (2008), using *grammatical evolution* with a fitness function based on a distance metric of the synthesized compositions to a pre-specified corpus of compositions.

### 3.1.3 Related Methods

Finally, this subsection presents a few examples that do not exactly use grammars, but utilize similar or borderline approaches.

The first one is the application of Kohonen's Dynamically Expanding Context (DEC) method to algorithmic composition (Kohonen et al., 1991). In DEC, a set of music examples is fed to the algorithm, which infers a model from the structure of the examples that may be construed as a stochastic context-sensitive grammar. The model is as parsimonious as possible, that is, the rules have as little contextual information as possible. Then, the inferred grammar is used to generate new compositions. Drewes and Högberg's (2007) work is also borderline, using regular tree grammars to generate a basic scaffold that is then modified by algebraic operations to generate a final music composition.

| Reference | Composition task | Comments |
|-----------|------------------|----------|
| Kohonen et al., 1991 | melody | Uses Kohonen's Dynamically Expanding Context |
| Drewes & Högberg, 2007 | generate variations on a melody | applies tree-based algebraic transformations |
| Cope, 1992 (EMI), 2000 (SARA, ALICE), 2005 (Emily Howell) | melody | EMI uses Augmented Transition Networks |

Table 4: References for Section 3.1.3, in order of appearance.

But the more famous example in this category is Cope's (1992) "Experiments in Musical Intelligence" (EMI), a software application able to analyze a set of musical compositions in a specific style (for example, Bach's) and to derive an Augmented Transition Network (ATN), i.e., a finite state automaton able to parse relatively complex languages. EMI then applies pattern-matching algorithms to extract signatures or short musical sequences characteristic of the style of the set of examples being analyzed, determining how and when to use these signatures in compositions with that style. After this analysis, the synthesis phase generates new music compositions that comply with the specifications encoded in the inferred ATN, with quite impressive results. He iterated EMI's design in other applications, like SARA and ALICE (Cope, 2000), but ultimately tried a new approach with yet another application, "Emily Howell". Cope (2005) reported that "Emily Howell" developed a unique style by a process of trial and error guided by human input; however, other researchers (Wiggins, 2008) have disputed the validity of his methodology.

## 3.2 Symbolic, Knowledge-Based Systems and Related Methods

Here, *knowledge-based system* is used as an umbrella term encompassing various rule-based systems under several different paradigms, with the common denominator of representing knowledge as more or less structured symbols. Since knowledge about musical composition has traditionally been structured as sets of more or less formalized rules for manipulating musical symbols (Anders & Miranda, 2011), knowledge-based and rule systems come as a natural way to implement algorithmic composition. In fact, it is extremely common for algorithmic composition systems to include some kind of composition rules at some point of the workflow. The most known early work on algorithmic composition is an example: classical rules for counterpoint were used in the generation of the first and second movements of the *Illiac Suite* (Hiller & Isaacson, 1958). Because of this, this subsection is mostly confined to the description of systems with strong foundations in AI (as expert systems), sidestepping to a certain degree the works of composers that are difficult to categorize, because of the *ad hoc* nature of their approaches and the very different language they use.

Starting with an exposition of early work, Gill's (1963) paper, already cited in Section 2.1, presented the first application of classical AI heuristics to algorithmic composition: he used a hierarchical search with backtracking to guide a set of compositional rules from Schoenberg's twelve-tone technique. Another notable example is Rothgeb's (1968) Ph.D. thesis: he encoded in SNOBOL a set of rules extracted from eighteenth century music treatises to harmonize the unfigured bass, that is to say, determine adequate chords from a sequence of bass notes.[14] He discovered that the classical rules were incomplete and incoherent to a certain extent.

| Reference | Composition task | Comments |
|---|---|---|
| Gill, 1963 | Schoenberg's twelve-tone technique | hierarchical search with backtracking |
| Rothgeb, 1968 | unfigured bass | implemented in SNOBOL |
| Thomas, 1985 (Vivace) | four-part harmonization | implemented in LISP |
| Thomas et al., 1989 (Cantabile) | Indian *raga* style | implemented in LISP |
| Steels, 1986 | four-part harmonization | uses Minsky's *frames* |
| Riecken, 1998 (Wolfgang) | melody | uses Minsky's SOM |
| Horowitz, 1995 | jazz improvisation | uses Minsky's SOM |
| Fry, 1984 (Flavors Band) | jazz improvisation and other styles | phrase processing networks (networks of agents encoding musical knowledge) |
| Gjerdingen, 1988 (Praeneste) | species counterpoint | implements a theory of how composers work |
| Schottstaedt, 1989 | species counterpoint | constraint-based search with backtracking |
| Löthe, 1999 | piano minuets | set of rules extracted from a classical textbook |
| Ulrich, 1977 | jazz improvisation | also uses a grammar (for jazz chords) |
| Levitt, 1981 | jazz improvisation | Criticized by Horowitz (1995) for being overly primitive |
| Hirata & Aoyagi, 1988 | jazz improvisation | uses logic programming |
| Rowe, 1992 (Cypher) | interactive jazz improvisation | uses Minsky's SOM |
| Walker, 1994 (ImprovisationBuilder) | interactive jazz improvisation | implemented in SmallTalk |
| Ames & Domino, 1992 (Cybernetic Composer) | jazz, rock | also uses Markov chains for rhythm |

Table 5: References for Section 3.2, in order of appearance.

These were recurring problems for many others implementing rules of composition straight from musical theory. For example, Thomas (1985) designed a rule-based system for four-part chorale harmonization implemented in Lisp[15], with the intent of clarifying the musical rules she taught to her students. Later, she designed another rule system (Thomas et al., 1989) for simple melody generation in the Indian *raga* style. Another example in harmonization is the use of Minsky's paradigm of *frames* by one of his students to encode a set of constraints to solve a relatively simple problem from tonal harmony, finding a passing chord between two others (Steels, 1979), and later to tackle the problem of four-part harmonization (Steels, 1986). Minsky developed other paradigms, such as *K-lines* and the

---

14. It should be noted that this stems from the practice of not completely specifying the harmonization, a problem that performers were expected to solve by improvisation.

15. All the systems discussed in this paragraph were implemented in Lisp.

| Reference | Composition task | Comments |
|---|---|---|
| Schwanauer, 1993 (MUSE) | four-part harmonization | presents the learning techniques in a similar way to Roads (1985, sect. 8.2) |
| Widmer, 1992 | harmonization | based on user evaluations of a training corpus |
| Spangler, 1999 | real-time four-part harmonization | prioritizes harmonic errors by severity, in order to refine the results |
| Morales & Morales, 1995 | species counterpoint | uses logic programming |

Table 6: References for Section 3.2.1, in order of appearance.

*Society of Mind* (SOM), which also influenced the work on algorithmic composition of two of his students, Riecken and Horowitz. Riecken (1998) used them in a system that composed monophonic melodies according to user-specified "emotional" criteria, while Horowitz (1995) used them in a system that improvised jazz solos. Fry's (1984) *phrase processing networks*, while not directly based on SOM, were specialized procedural representations of networks of agents implementing musical transformations to encode knowledge about jazz improvisation and other styles.

Other researchers have also explored different ways to generate species counterpoint with rule-based systems: Gjerdingen (1988) implemented a system based on the use of several pre-specified musical schemata, implementing a theory of how composers work, while Schottstaedt (1989) used a more formal approach: a constraint-based search with backtracking. He followed a classical rulebook on species counterpoint, to the point of bending some rules and creating new ones in order to get as close as possible to the scores serving as examples in that book. Also on the formal side, Löthe (1999) extracted a set of rules from a classical textbook for composing minuets.

Other music styles demanded different approaches: as jazz performances are improvisations over existing melodies, knowledge-based systems for jazz were structured as more or less sophisticated analysis-synthesis engines. For example, the work of Ulrich (1977): his system analyzed a melody and fitted it to a harmonic structure. Another student of Minsky (Levitt, 1981) implemented a rule-based jazz improviser formulating some of the rules as constraints, while Hirata and Aoyagi (1988) encoded the rules in logic programming, trying to design a more flexible system. Rowe (1992) used a SOM architecture[16] for Cypher, an analysis-synthesis engine able to play jazz interactively with a human performer, notable for its flexibility and the musical knowledge encoded into it. Also, Walker (1994) implemented an object-oriented analysis-synthesis engine able to play jazz interactively with a human performer, and Ames and Domino (1992) implemented a hybrid system (using rules and Markov chains) for the generation of music in several popular genres.

### 3.2.1 Rule Learning

While the knowledge implemented in rule-based systems is usually static, part of the knowledge may be dynamically changed or *learned*. The natural term for this concept is *machine learning*, but its meaning is unfortunately vague, because it is used as a catch-all for many methods, including neural networks and Markov chains.

---

16. He was not a student of Minsky, though.

A few examples of rule-based learning systems have been developed. For example, Schwanauer (1993) implemented MUSE, a rule-based system for solving several tasks in four-part harmonization. While the core ruleset was static, a series of constraints and directives for the composition process where also built in the system, and their application was also used to dynamically change the rule priorities. Additionally, when the system successfully solved a task, it was able to deduce new composite rules by extracting patterns of rule application. Widmer (1992) implemented another example: a system for the harmonization of simple melodies. It was based on user evaluations of a training corpus: from a hierarchical analysis of the training melodies and theirs evaluations, it extracted rules of harmonization. Spangler (1999) implemented a system for generating rule systems for harmonizing four-part chorales in the style of Bach, with the constraint of doing the harmonization in real time. The rulesets were generated by analyzing databases of examples with algorithms that applied formal concepts of information theory for distilling the rules, and the violations of harmonic rules were prioritized in order to refine the results. Using the framework of logic programming, Morales and Morales (1995) designed a system that learned rules of classical counterpoint from musical examples and rule templates.

### 3.2.2 Rule-Based Methods and Evolutionary Algorithms

The most intuitive way to hybridize rule-based knowledge systems and evolutionary algorithms is to craft a fitness function from the ruleset. This can be done efficiently for domains whose rules have been adequately codified, and compliance with the rules can be expressed as a graduated scale, instead of a binary (yes/no) compliance.

A good example is four-part baroque harmonization for a pre-specified melody, which lends itself particularly well to this approach. McIntyre (1994) extracted a set of rules for performing this harmonization from classical works, and codified them as a set of scoring functions. The fitness was a weighted sum of these scores, with a tiered structure: some scores were not added unless other specific scores had values above some thresholds (because they were more critical or prerequisites to produce good harmonizations). A slightly different approach was used by Horner and Ayers (1995): they defined two classes of rules: one for defining acceptable voicings for individual chords, used to enumerate all possible voicings, and another for defining how the voices are allowed to change between successive chords. An evolutionary algorithm was used to find music compositions, whose search space was constructed with the enumeration of voicings (first class of rules). The fitness of each candidate solution was simply the amount of violated rules from the second class. Phon-Amnuaisuk et al. (1999) also did four-part harmonization using a set of rules to build the fitness function and musical knowledge to design the genotype and the mutation and crossover operators, but the lack of global considerations in the fitness function led to modest results. In contrast, Maddox and Otten (2000) got good results implementing a system very similar to McIntyre's (1994), but using a more flexible representation, resulting in a larger search space of possible individuals, and without the tiered structure in the fitness function, enabling a less constrained search process.

Another good example is species counterpoint: Polito et al. (1997) extracted rules for species counterpoint from a classic eighteenth century music treatise, using them to define fitness functions in a multi-agent genetic programming system: each agent performed a set of composition or transformation operations on a given melody specified as a seed, and they cooperated to produce the composition. Gwee (2002) exhaustively studied the computational complexity of problems related to the generation of species counterpoint with rulesets, and implemented an evolutionary algorithm whose fitness function was based on a set of fuzzy rules (although he also experimented with trained artificial neural networks as fitness functions).

| Reference | Composition task | Comments |
|---|---|---|
| McIntyre, 1994 | four-part harmonization | explores several schemes to combine rules into the fitness function |
| Horner & Ayers, 1995 | four-part harmonization | two stages: enumeration of possible chord voicings, evolutionary algorithm for voice-leading rules |
| Phon-Amnuaisuk et al., 1999 | four-part harmonization | criticizes vanilla evolutionary algorithms for generating unstructured harmonizations |
| Maddox & Otten, 2000 | four-part harmonization | similar to McIntyre's (1994) |
| Polito et al., 1997 | species counterpoint | multi-agent genetic programming system |
| Gwee, 2002 | species counterpoint | fuzzy rules |

Table 7: References for Section 3.2.2, in order of appearance.

### 3.2.3 Constraint Satisfaction

Gradually (in a process that spanned the 1980s and 1990s), some researchers on algorithmic composition with rule-based systems adopted formal techniques based on logic programming. For example, Boenn et al. (2008) used *answer set programming* to encode rules for melodic composition and harmonization. However, most of the work on logic programming has been under a different paradigm: the formulation of algorithmic composition tasks as *constraint satisfaction problems* (CSPs). Previously referenced work, as Steels's (1979), Levitt's (1981), Schottstaedt's (1989) and Löthe's (1999) can be seen as part of a gradual trend towards the formulation of musical problems as CSPs[17], although *constraint logic programming* (CLP) came to be the tool of choice to solve CSPs. Good surveys on CLP for algorithmic composition have been written by Pachet and Roy (2001) and Anders and Miranda (2011).

Ebcioğlu worked for many years in this area, achieving notable results. In a first work implemented in Lisp (Ebcioğlu, 1980), he translated rules of fifth-species strict counterpoint to composable Boolean functions (he had to add rules of his own to bring the system into producing acceptable results, though), and used an algorithm that produced an exhaustive enumeration of the compositions satisfying a previously arranged set of rules: basically, he implemented a custom engine for logic programming in Lisp. Over the next decade, he the tackled the problem of writing four-part chorales in the style of J. S. Bach. Finally, he produced CHORAL, a monumental expert system (Ebcioğlu, 1988), distilling into it 350 rules to guide the harmonization process and the melody generation. To keep the problem tractable, he designed a custom logic language (BSL) with optimizations over standard logic languages, as backjumping. His system received substantial publicity, and was supposed to reach the level of a talented music student, in his own words.

---

17. While Gill's (1963) implementation was formulated as a CSP, it was somewhat primitive by later standards.

Following Ebcioğlu's work, many constraint systems have been implemented for harmonization or counterpoint. Tsang and Aitken (1991) implemented a CLP system using Prolog to harmonize four-part chorales. However, their system was grossly inefficient.[18] Ovans and Davison (1992) described an interactive CSP system for first-species counterpoint, where a human user drove the search process, and the system constrained the possible outputs (according to counterpoint rules) as the search progressed. They took care of efficiency by using arc-consistency in the resolution of the constraints. Ramírez and Peralta (1998) solved a different problem: given a monophonic melody, their CLP system generated a chord sequence to harmonize it. Phon-Amnuaisuk (2002) implemented a constraint system for harmonizing chorales in the style of J. S. Bach, but with an innovation over previous systems: to add knowledge to the system about how to apply the rules and control the harmonization process explicitly, thus modulating the search process in an explicit and flexible way. Anders and Miranda (2009) analyzed a Schoenberg's textbook on the theory of harmony, programming a system in Strasheela (see below) to produce self-contained harmonic progressions, instead of harmonizing pre-existing melodies, as most other constraint systems do.

While many CLP systems have been implemented to solve classical problems in harmonization or counterpoint, some researchers have studied the application of CLP techniques to different problems. In a very simple application, Wiggins (1998) used a CLP system to generate short fragments of serial music. Zimmermann (2001) described a two-stage method, where both stages used CLP: the first stage (AARON) took as input a "storyboard" to specify the mood of a composition as a function of time, and generated a harmonic progression and a sequence of directives. The second (COMPOzE) generated a four-part harmonization according to the previously arranged progression and directives; the result was intended as background music. Laurson and Kuuskankare (2000) studied constraints for the instrumentation[19] of guitars and trumpets (i.e., constraints for composing music easily playable in these instruments). Chemillier and Truchet (2001) analyzed two CSPs: a style of Central African harp music, and Ligeti textures. They used heuristic search in their analyzes instead of backtracking, heralding OMClouds' approach to constraint programming (see below). Sandred (2004) proposed the application of constraint programming to rhythm.

Several general-purpose constraint programming systems for algorithmic composition have been proposed (i.e., languages and environments to program the constraints). One of the earliest examples was Courtot's (1990) CARLA, a CLP system for generating polyphonies with a visual front-end and a rich, extendable type system designed to represent relationships between different musical concepts. Pachet and Roy (1995) implemented another general-purpose musical CLP (Backtalk) in an object-oriented framework (MusES), designing a generator of four-part harmonizations on top of it. Their key contribution was a hierarchical arrangement of constraints on notes and chords, dramatically decreasing the (both cognitive and computational) complexity of the resulting constraint system.

---

18. In spite of using just 20 rules, it required up to 70 megabytes of memory to harmonize a phrase of 11 notes.

19. That is to say, take into account the way an instrument is played when composing its part.

| Reference | Composition task | Comments |
|---|---|---|
| Boenn et al., 2008 | melody and harmonization | answer set programming |
| Ebcioğlu, 1980 | species counterpoint | implemented in LISP |
| Ebcioğlu, 1988 (CHORAL) | four-part harmonization | implemented in a custom logic language (BSL) |
| Tsang & Aitken, 1991 | four-part harmonization | very inefficient |
| Ovans & Davison, 1992 | species counterpoint | interactive search |
| Ramírez & Peralta, 1998 | melody harmonization | simpler constraint solver |
| Phon-Amnuaisuk, 2002 | four-part harmonization | explicit control over the search process |
| Anders & Miranda, 2009 | Schoenberg's Theory of Harmony | implemented in Strasheela |
| Wiggins, 1998 | Schoenberg's twelve-tone technique | very simple demonstration |
| Zimmermann, 2001 (Coppelia) | structure, melody, harmonization, rhythm | two stages: harmonic plan (Aaron) and execution (Compoze) |
| Laurson & Kuuskankare, 2000 | guitar and trumpet instrumentation | implemented with PWConstraints |
| Chemillier & Truchet, 2001 | African harp and Ligeti textures | implemented in OpenMusic |
| Sandred, 2004 | rhythm | implemented in OpenMusic |
| Courtot, 1990 (CARLA) | polyphony, general purpose | early general-purpose system |
| Pachet & Roy, 1995 (BackTalk) | four-part harmonization | implemented in MusEs |
| Rueda et al., 1998 | polyphony, general purpose | describes PWConstraints (implemented in PatchWork) and Situation (implemented in OpenMusic) |
| Rueda et al., 2001 | general purpose | describes PiCO |
| Olarte et al., 2009 | general purpose | describes *ntcc* |
| Allombert et al., 2006 | interactive improvisation | uses *ntcc* |
| Rueda et al., 2006 | interactive improvisation | uses *ntcc* and Markovian models |
| Pachet et al., 2011 | melody | integrates Markovian models and constraints |
| Truchet et al., 2003 | general purpose | describes OMClouds |
| Anders, 2007 | general purpose | describes Strasheela |
| Sandred, 2010 | general purpose | describes PWMC. Implemented in PatchWork |
| Carpentier & Bresson, 2010 | orchestration | uses multi-objective optimization to discover candidate solutions. Interfaces with OpenMusic and MAX |
| Yilmaz & Telatar, 2010 | harmonization | fuzzy logic |
| Aguilera et al., 2010 | species counterpoint | probabilistic logic |
| Geis & Middendorf, 2008 | four-part harmonization | multi-objective Ant Colony Optimization |
| Herremans & Sorensena, 2012 | species counterpoint | variable neighborhood with tabu search |
| Davismoon & Eccles, 2010 | melody, rhythm | uses simulated annealing to combine constraints with Markov processes |
| Martin et al., 2012 | interactive improvisation | implemented in MAX |

Table 8: References for Section 3.2.3, in order of appearance.

Rueda et al. (1998) reviewed two other early general-purpose systems, PWConstraints and Situation. PWConstraints was able to (relatively easily) handle problems in polyphonic composition through a subsystem (score-PMC), while Situation was more flexible and implemented more optimizations in its search procedures. PiCO (Rueda et al., 2001) was an experimental language for music composition that seamlessly integrated constraints, object-oriented programming and a calculus for concurrent processes. The idea was to use constraint programming to specify the voices in a composition, and to use the concurrent calculus to harmonize them. The authors also implemented a visual front-end to PiCO for ease of use, Cordial. In a similar way to PiCO, *ntcc* was another language for constraint programming that implemented primitives for defining concurrent systems, although it was not specifically designed for algorithmic composition. *ntcc* has been proposed to generate rhythm patterns and as a more expressive alternative to PiCO (Olarte et al., 2009), and has mainly been used for machine improvisation: Allombert et al. (2006) used it as the improvisation stage of their two-stage system (the first stage used a temporal logic system to compose abstract temporal relationships between musical objects, while the *ntcc* stage generated concrete music realizations), and Rueda et al. (2006) used *ntcc* to implement a real-time system that learned a Markovian model (using a Factor Oracle) from musicians and concurrently applied it to generate improvisations. Not related to *ntcc*, Pachet et al. (2011) has also proposed a framework to combine constraint satisfaction and Markov processes.

OMClouds (Truchet et al., 2003) was another general-purpose (but purely visual) constraint system for composition, but its implementation set it apart from most other formal systems: internally, the constraints are translated to cost functions. Instead of the optimized tree search with backtracking usual in CLP, an adaptive tabu search was performed, seeking to minimize a solution with minimal cost. This avoids some problems inherent to constraint programming, such as overconstraining, but it cannot be guaranteed to completely navigate the search space. Anders (2007) implemented Strasheela, a system that was expressly designed to be highly flexible and programmable, aiming to overcome a perceived limitation of previous general-purpose systems: the difficulty to implement complex with constraints related to multiple aspects of the compositions process. Finally, another purely visual constraint system, PWMC, was proposed by Sandred (2010) to overcome perceived limitations of score-PMC. It was able to handle constraints concerning not only pitch structure as score-PMC, but also rhythm and metric structure.

It should be stressed that, while CLP has become the tool of choice to solve CSPs, other approaches are also used. Previously cited OMClouds is just one of these. Carpentier and Bresson (2010) implemented a mixed system for orchestration that worked in a curious way: the user fed the system with a target sound and a set of symbolic constraints; a multi-objective evolutionary algorithm found a set of orchestration solutions matching the target sound, and a local search algorithm filtered out the solutions not complying with the constraints. Yilmaz and Telatar (2010) implemented a system for simple constraint harmonization with fuzzy logic, while Aguilera et al. (2010) used probabilistic logic to solve first-species counterpoint. More exotic solutions have been proposed, as the use of *Ant Colony Optimization* with a multi-objective approach to solve the constraints of Baroque harmonization (Geis & Middendorf, 2008), *variable neighborhood with tabu search* to solve soft constraints for first-species counterpoint (Herremans & Sorensena, 2012), or *simulated*

| Reference | Composition task | Comments |
|-----------|------------------|----------|
| Pereira et al., 1997 | Baroque music | hierarchical analysis and representation |
| Ribeiro et al., 2001 (MuzaCazUza) | Baroque music | generates a melody from a harmonic line |
| Ramalho & Ganascia, 1994 | jazz improvisation | uses a rule-based system for analysis |
| Parikh, 2003 | jazz improvisation | also uses a rule system to analize music |
| Eigenfeldt & Pasquier, 2010 | jazz chord progressions | also uses Markovian models |
| Sabater et al., 1998 | harmonization | also uses a rule system |

Table 9: References for Section 3.2.4, in order of appearance.

*annealing* to combine constraints with Markov processes (Davismoon & Eccles, 2010). Finally, Martin et al. (2012) presented an even more exotic approach: a real-time music performer that reacted to its environment. While some of the aspects of the music where controlled by Markov chains, others where expressed as a CSP. To solve this CSP in real time, a solution was calculated at random (but quickly) using binary decision diagrams.

### 3.2.4 Case-Based Reasoning

Case-based reasoning (CBR) is another formal framework for rule-based systems. In the CBR paradigm, the system has a database of cases, that can be defined as instances of a problem with their corresponding solutions. Usually, a case also contains structured knowledge about how the problem is solved in that case. When faced with a new problem, the system matches it against the case database. Unless the new problem is identical to one recorded in the case database, the system will have to select a case similar to the new problem, and adapt the corresponding solution to the new problem. If the new solution is deemed appropriate, a new case (recording the new problem along with the new solution) may be included in the database.

Several researchers have used CBR for algorithmic composition. Pereira et al. (1997) implemented a system that generated its case database from just three Baroque music pieces, which were analyzed into hierarchical structures; the cases were their nodes. The system composed just the soprano melodic line of the piece, searching for similar cases in its case database. The results were comparable to the output of a first-year student, according to music experts consulted by the authors. An intersecting set of researchers implemented a simpler CBR composing system (Ribeiro et al., 2001) that generated a melody from a harmonic line, this time with a case database generated from just six Baroque pieces. Cases were represented in a different way, tough: each case represented the rhythm, the melody and other attributes associated to a chord in a given context. To generate a new music piece, a harmonic line was specified, and the system fleshed out the music piece by matching the cases to the harmonic line.

Hybrid systems have also been proposed. Ramalho and Ganascia (1994) proposed a jazz improviser that used a rule system to analyze incoming events (for example, the ongoing sequence of chords) and a CBR engine to improvise. The case database was assembled by extracting patterns from transcriptions of jazz recordings, and consisted of descriptions of contexts and how to play in these contexts. During the improvisation, the current context was analyzed by the rule system, and those cases applying in the current context were extracted from the database and combined to determine the output of the improviser. Considering that Ramalho and Ganascia's system was too inflexible, Parikh (2003) implemented another jazz improviser, intending to use a large case database containing jazz fragments from various sources, in order to get a system with a style of its own. Eigenfeldt and Pasquier (2010) used a case-based system to generate variable-order Markov models for jazz chord progressions.

| Reference | Composition task | Comments |
|-----------|------------------|----------|
| Haus & Sametti, 1991 (Scoresynth) | melody | Petri nets |
| Lyon, 1995 | melody | Petri nets to encode Markov chains |
| Holm, 1990 | melody, sound synthesis | inspired in CSP process algebra |
| Ross, 1995 (MWSCCS) | melody | custom process algebra |
| Rueda et al., 2001 | general purpose | describes PiCO |
| Allombert et al. (2006) | interactive improvisation | uses *ntcc* |
| Rueda et al., 2006 | interactive improvisation | uses *ntcc* and Markovian models |
| (Olarte et al., 2009) | general purpose, rhythms | describes *ntcc* |

Table 10: References for Section 3.2.5, in order of appearance.

Outside the jazz domain, a hybrid system for harmonizing melodies of popular songs was implemented by Sabater et al. (1998): given a melody, the system sequentially decided the chords to harmonize it. If the CBR module failed to match a case, the system fell back to a simple heuristic rule system to select an appropriate chord. As the harmonized output was added to the case database, the CBR module gradually learned over time from the rule system.

### 3.2.5 Concurrency Models

Concurrency models can be described as formal languages to specify, model and/or reason about distributed systems. They provide primitives to precisely define the semantics of interaction and synchronization between several entities. Their main application has been modeling and designing distributed or concurrent computer systems, but they have also been used as languages to partially or fully model the composition process, because music composition can be formulated as an endeavor to carefully synchronize streams of music events produced by several interacting entities. Concerning algorithmic composition, the most used concurrency models have been Petri nets and several kinds of process algebras, also known as process calculi. Detailed descriptions of these models are beyond the scope of this survey; see for example Reisig's (1998) book on Petri nets and Baeten's (2005) survey on process algebras for more information.

Petri nets have been used as the basis for Scoresynth (Haus & Sametti, 1991), a visual framework for algorithmic composition in which Petri nets were used to describe transformations of musical objects (sequences of notes and musical attributes), and the synchronization between musical objects was implicit in the structure of the net. Petri nets have also been used as an efficient and compact way to implement Markov chains (Lyon, 1995).

Process algebras were first used for algorithmic composition by Holm (1990), although his model (inspired by Hoare's algebra, CSP) was more geared towards sound synthesis than music composition. A more proper example is Ross's (1995) MWSCCS, an extension (adding concepts for music composition) of a previously existing algebra (WSCCS). Specifications for algorithmic composition written in MWSCCS were meant to resemble grammatical specifications, but with a richer expressive power. Later examples have also been cited in Section 3.2.3, as the PiCO language (Rueda et al., 2001), which integrated logical constraints and a process algebra. Also cited in that Section, *ntcc* is a process algebra that has been used to implement machine improvisation (Allombert et al., 2006) and to drive a Markovian model (using a Factor Oracle) for real-time machine learning and improvisation (Rueda et al., 2006). It has also been proposed to generate rhythm patterns, and as a more expressive alternative to PiCO (Olarte et al., 2009).

## 3.3 Markov Chains and Related Methods

Conceptually, a Markov chain is a simple idea: a stochastic process, transiting in discrete time steps through a finite (or at most countable) set of states, *without memory*: the next state depends just on the current state, not on the sequence of states that preceded it or on the time step. In their simplest incarnations, Markov chains can be represented as labeled directed graphs: nodes represent states, edges represent possible transitions, and edge weights represent the probability of transition between states. However, Markov chains are more commonly represented as probability matrices.

When Markov chains are applied to music composition, the probability matrices may be either induced from a corpus of pre-existing compositions (*training*), or derived by hand from music theory or by trial-and-error. The former is the most common way to use them in research, while the latter is more used in software tools for composers. An important design decision is how to map the states of the Markov chain to musical objects. The simplest (but fairly common) mapping just assigns a sequential group of notes to each state, with the choice of just one note (instead of a larger sequence) being fairly common.

It is also common to extend the consideration of the "current state": in an $n$-th order Markov chain, the next state depends on the last $n$ states, not just the last one. As a consequence, the probability matrix has $n + 1$ dimensions. In algorithmic composition, Markov chains are mostly used as generative devices (generating a sequence of states), but they can also be used as analysis tools (evaluating the probability of a sequence of states). In the latter case, the term $n$-*gram* is also used, though strictly speaking it refers to a sequence of $N$ states.

Markov chains were a very popular method in the early years of algorithmic composition. Early examples have already been reviewed in Section 2.1; additionally, Ames (1989) also provides a good survey. However, Markov chains generated from a corpus of pre-existing compositions captured just local statistical similarities, and their limitations soon became apparent (Moorer, 1972): low-$n$ Markov chains produced strange, unmusical compositions that wandered aimlessly, while high-$n$ ones essentially rehashed musical segments from the corpus and were also very computationally expensive to train.

| Reference | Composition task | Comments |
|---|---|---|
| Tipei, 1975 | melody | Markov chains are part of a larger, *ad hoc* system |
| Jones, 1981 | melody | very simple introduction for composers |
| Langston, 1989 | melody | dynamic weights |
| North, 1991 | melody | Markov chains are part of a larger, *ad hoc* system |
| Ames & Domino, 1992 (Cybernetic Composer) | jazz, rock | uses Markov chains for rhythms |
| Visell, 2004 | real-time generative art installation | Liberal use of the concept of HMM. Implemented in MAX |
| Zicarelli, 1987 (M and Jam Factory) | interactive improvisation | commercial GUI applications |
| Ariza, 2006 | | alternative representation for transition matrices. Implemented in athenaCL |
| Ponsford et al., 1999 | composing *sarabande* pieces | adds symbols to expose the structure of the pieces |
| Lyon, 1995 | melody | implements Markov chains with Petri nets |
| Verbeurgt et al., 2004 | melody | two stages: a Markovian model and an artificial neural network |
| Thom, 2000 (BoB) | interactive jazz improvisation | statistical machine learning |
| Lo & Lucas, 2006 | melody | evolutionary algorithm, Markov chains used in the fitness function |
| Werner & Todd, 1997 | melody | co-evolutionary algorithm, Markov chains used as evolvable fitness functions |
| Thornton, 2009 | melody | grammar-like hierarchy of Markov models |
| Cruz-Alcázar and Vidal-Ruiz (1998) | melody | analysis with grammatical inference, generation with Markovian models |
| Gillick et al. (2009) | jazz improvisation | analysis with grammatical inference, generation with Markovian models |
| Eigenfeldt & Pasquier, 2010 | jazz chord progressions | uses case-based reasoning |
| Davismoon & Eccles, 2010 | melody, rhythm | uses simulated annealing to combine constraints with Markov processes |
| Pachet et al., 2011 | melody | integrates Markovian models and constraints |
| Grachten, 2001 | jazz improvisation | integrates Markovian models and constraints |
| Manaris et al., 2011 | interactive melody improvisation | Markov chains generate candidates for an evolutionary algorithm |
| Wooller & Brown, 2005 | transitioning between two melodies | alternates Markov chains from the two melodies |

Table 11: References for Section 3.3, in order of appearance.

Because of this, Markov chains came to be seen as a source of raw material, instead of a method to truly compose music in an automated way, except for some specialized tasks such as rhythm selection (McAlpine et al., 1999). Therefore, while research interest in Markov chains receded in subsequent years as these limitations became apparent and other methods were developed, they remained popular among composers. However, citing even a relevant subset of all the works were composers use Markov chains as part of their compositional process would inflate the reference list beyond reasonable length. A few typical examples of Markov chains used by composers (sometimes as part of a larger automatic compositional framework or software system) are the papers of Tipei (1975), Jones (1981), Langston (1989, although he used dynamically computed weights), North (1991) and Ames and Domino (1992). It should be noted that composers sometimes deconstruct formal methods to adapt them to their own purposes, as when Visell (2004) used the concept of a Hidden Markov Model (described below) to implement a manually-tuned real-time generative art system.

In addition, many software suites use Markov chains to provide musical ideas to composers, even if the probabilities are specified by hand instead of generated from a corpus. Ariza (2006) gives a compact list of software suites and experimental programs using Markov chains, while his thesis (Ariza, 2005b) provides a comprehensive view of the field. Much has been done for usability in this field, by using GUI interfaces (Zicarelli, 1987), but also developing more effective ways to encode the probability matrices, for example as compact string specifications (Ariza, 2006).

However, novel research on Markov chains for algorithmic compositions has still been carried out in several ways. For example, Ponsford et al. (1999) used a corpus of *sarabande* pieces (relatively simple dance music) to generate new compositions using Markov models[20], but with a pre-processing stage to automatically annotate the compositions of the corpus with symbols to make explicit their structure, and a post-processing stage using a template to constrain the structure of the synthesized composition, in order to generate minimally acceptable results. Another way is the hybridization of Markov chains with other methods. For example, Lyon (1995) used Petri nets as an efficient and compact way to implement Markov chains, while Verbeurgt et al. (2004) used Markov chains[21] to generate a basic pattern for the melody, which was then refined with an artificial neural network. In the BoB system (Thom, 2000), Markov chains were trained by statistical learning: from a set of jazz solos, statistical signatures were extracted for pitches, melodic intervals and contours. Then, these signatures were used to define the transition probabilities of a Markov chain whose output was sampled to generate acceptable solos. Lo and Lucas (2006) trained Markov chains with classic music pieces, but, instead of generating compositions with them, used them as fitness evaluators in an evolutionary algorithm to evolve melodies encoded as sequences of pitches. Werner and Todd (1997) also used Markov chains to evaluate simple (32-note) melodies, but with the particularity that the chains themselves were also subject to evolution, to investigate sexual evolutionary dynamics. Thornton (2009) defined a set of grammar-like rules from an existing composition, inferring a hierarchy of Markov models to use statistical patterns of the analyzed composition at multiple levels. As already mentioned in Section 3.1, Cruz-Alcázar and Vidal-Ruiz (1998) and Gillick et al. (2009) used grammatical inference with Markovian models. Regarding symbolic methods, Eigenfeldt and Pasquier (2010) used a case-based system to generate Markov processes for jazz chord progressions, (Davismoon & Eccles, 2010) used *simulated annealing* to combine constraints and Markov processes, and Pachet et al. (2011) proposed a framework to combine Markovian generation of music with rules (constraints) to produce better results.

---

20. Their work is commonly cited in the literature as grammatical, but their methodology is thoroughly statistical.
21. Also reviewed in Section 3.4.

Additionally, Markov chains remained a feasible option for restricted problems (for example, real-time performances, as jazz improvisation), as their limitations were less apparent in these cases than in the generation of whole compositions. For example, Grachten (2001) developed a jazz improviser where Markov chains generated duration and pitches, and then a system of constraints refined the output, and pre-defined *licks* (short musical patterns) were inserted at appropriate times. Manaris et al. (2011) also implemented an improviser, using a Markov model with user input to generate a population of candidate melodies, feeding them into an evolutionary algorithm, whose fitness function rewarded melodies whose metrics were similar to the user input's metrics. A different (but also restricted) problem was studied by Wooller and Brown (2005): applying Markov chains to generate musical transitions (*morphings*) between two different pieces in a simple application of non-linear music, by stochastically alternating between two Markov chains, each one trained with one of the pieces.

### 3.3.1 RELATED METHODS

More sophisticated Markovian models (and related statistical methods; see the survey in Conklin, 2003) have also been applied for algorithmic composition, as in Pachet's (2002) *Continuator*, a real-time interactive music system. The *Continuator* departs from common Markov chain implementations in that it uses *variable-order* (also known as *mixed-order*) Markov chains[22], which are not constrained to a fixed $n$ value, and can be used to get the best of low and high-$n$ chains. Conklin and Witten (1995) implemented a sophisticated variable-order scheme[23], whose main feature was the consideration in parallel of *multiple viewpoints* or sequences of events in the compositions (for example, pitches, durations, contours, etc.), instead of integrating them all in a unique sequence of symbols, as it was common for most implementations of Markov chains. Variable-order Markov chains have also been used as part of a larger real-time music accompaniment system (Martin et al., 2012). Other variable-order schemes used in algorithmic composition, formulated in a machine learning framework, are *Prediction Suffix Trees* (PSTs, Dubnov et al., 2003), more space-efficient structures like *Factor Oracles*[24] (Assayag & Dubnov, 2004), and *Multiattribute Prediction Suffix Graphs* (MPSGs, Triviño Rodríguez & Morales-Bueno, 2001), which can be considered an extension of PSTs to consider *multiple viewpoints* as in Conklin and Witten's work. Sastry (2011) also used multiple viewpoints and PSTs to modelize Indian *tabla* compositions, though his model could also be used to generate new compositions.

Hidden Markov Models (HMMs) also are generalizations of Markov chains that have been used for algorithmic composition. A HMM is a Markov chain whose state is unobservable, but some state-dependent output is visible. Training a HMM involves not only determining a matrix of transition probabilities, but also a matrix of output probabilities (that is, for each state, the probability of each possible output). Then, given a sequence of outputs, it is possible to compute the most likely sequence of states to produce that sequence of outputs, using the Viterbi dynamic programming algorithm. In this way, HMMs find a globally optimized sequence of states, while simpler Markov methods perform just local optimization. When applied to algorithmic composition, HMMs are appropriate to add elements to an existing composition (most commonly, counterpoint and harmonization), given a set of pre-existing examples: the composition is modeled as a sequence of outputs of the HMM, and the additions are computed as the most likely sequence of states of the HMM.

---

22. It should be noted that Kohonen's method (Kohonen et al., 1991), reviewed in Section 3.1.3, is similar (in some ways) to variable-order chains.

23. Conklin and Witten's method has also been described as grammatical, but they are included here because their emphasis in formal statistical analysis.

24. Also implemented with a concurrent constraint paradigm by Rueda et al. (2006). See Section 3.2.3 and Section 3.2.5 for more details.

| Reference | Composition task | Comments |
|---|---|---|
| Pachet, 2002 (Continuator) | interactive improvisation | variable-order |
| Conklin & Witten, 1995 | Bach chorales | multiple viewpoint systems |
| Martin et al., 2012 | interactive improvisation | variable-order; implemented in MAX |
| Dubnov et al., 2003 | melody | Prediction Suffix Trees. Implemented in OpenMusic |
| Rueda et al., 2006 | interactive improvisation | uses *ntcc* and Factor Oracles |
| Assayag & Dubnov, 2004 | melody | Factor Oracles. Implemented in OpenMusic |
| Triviño Rodríguez & Morales-Bueno, 2001 | melody | Multiattribute Prediction Suffix Graphs |
| Sastry, 2011 | improvisation of *tabla* rhythms | multiple viewpoints and Prediction Suffix Trees. Implemented in MAX |
| Farbood & Schoner, 2001 | species counterpoint | Hidden Markov Models |
| Biyikoglu, 2003 | four-part harmonization | Hidden Markov Models |
| Allan, 2002 | four-part harmonization | Hidden Markov Models |
| Morris et al., 2008 (SongSmith) | melody harmonization | Hidden Markov Models |
| Schulze, 2009 (SuperWillow) | melody, rhythm, two-voice harmonization | Hidden Markov Models and Prediction Suffix Trees |
| Yi & Goldsmith, 2007 | four-part harmonization | Markov Decision Processes |
| Martin et al., 2010 | interactive improvisation | Partially Observable Markov Decision Processes |

Table 12: References for Section 3.3.1, in order of appearance.

Farbood and Schoner (2001) implemented the earliest example of a HMM for algorithmic composition: they trained a second-order HMM to generate Palestrina-style *first-species counterpoint* (the simplest way to write counterpoint), defining the training set from rules used to teach counterpoint. A related problem is to train HMMs with a set of chorale harmonizations in the style of J.S. Bach in order to get more Bach-like harmonizations. This problem has been researched by Biyikoglu (2003) and Allan (2002); the latter divided the problem of harmonization into the same three subtasks as in HARMONET (Hild et al., 1992). For Microsoft's *SongSmith* software, Morris et al. (2008) trained a HMM with 300 *lead sheets* (specifications for song melodies) to generate chords to accompany a user-specified vocal melody, parametrizing the resulting system with a very intuitive interface for non-technical users. Schulze (2009) generated music in several styles using mixed-order Markov chains to generate the melodies, and HMMs to harmonize them.

Markov Decision Processes (MDPs) are another generalization of Markov models, in which an agent maximizes some utility function by taking actions to probabilistically influence the next state, and Partially Observable MDPs (POMDPs) represent the corresponding generalization of HMMs. Experimental systems for algorithmic composition have been implemented with MDPs (Yi & Goldsmith, 2007) and POMDPs (Martin et al., 2010), though it is not clear that these sophisticated models offer definitive advantages over simpler ones.

### 3.4 Artificial Neural Networks and Related Methods

Artificial Neural Networks (ANNs) are computational models inspired in biological neural networks, consisting of interconnected sets of *artificial neurons*: very simple computational devices that aggregate numeric inputs into a single numeric output using a (generally) simple but nonlinear function. Some neurons have connections that are set externally (input connections), while other have output signals intended to be read as the result of the network's computation (output connections). Typically, neurons are organized in recurrent networks (some or all neurons have inputs that come from other neurons) with several interconnected *layers*, and many variations can be found in the literature. ANNs are typically used as a machine learning method, using a set of *examples* (input patterns) to *train* the network (i.e., to set the weights of the connections between neurons), in order to use it to recognize or generate similar patterns. Effectively, this means that neural networks need a pre-existing corpus of music compositions (all of them in a very similar style, generally); therefore they can at most *imitate* the style of the training examples. Most papers use a *supervised learning* approach, meaning that the examples in the training set are associated with a signal, and the ANN learns this association. An important aspect of ANN design is the modelization of musical composition, that is, the mapping between music or music notation and the inputs and outputs of the network. Another important aspect is the way in which compositions are fed to the ANNs: they may be presented as temporal patterns in the network inputs, which are usually windowed in segments, but in some cases they are fed at once (as wholes) to the ANNs (these implementations do not scale well, though, because of the big ANNs needed to model long compositions).

ANNs were first used during the 1970s and 1980s to analyze musical compositions, creating artificial models of cognitive theories of music (Todd & Loy, 1991), but they were later adapted for music composition. The first example was implemented by Todd (1989), who used a three-layered recurrent ANN designed to produce a temporal sequence of outputs encoding a monophonic melody, each output signal of the network representing an absolute pitch. Given a set of one or more composition examples, the ANN was trained to associate a single input configuration to the output temporal sequence of the corresponding composition. Then, feeding input configurations different to the ones used during the training created melodies interpolated between the ones used during the training. If just one melody was used during the training, the result was an extrapolation from it. Later that year, Duff (1989) published another early example, but using a different approach, encoding relative

| Reference | Composition task | Comments |
|---|---|---|
| Todd, 1989 | melody | three layers, recurrent |
| Duff, 1989 | melody | two layers, recurrent |
| Mozer, 1991 (CONCERT) | melody | psychologically-grounded representation of pitch |
| Lewis, 1991 | | feedforward model, used as the fitness function in an optimization algorithm |
| Shibata, 1991 | harmonization | feedforward model |
| Bellgard & Tsang, 1992 | harmonization | effective Boltzmann machine |
| Melo, 1998 | harmonization | the ANN is trained to model music tension |
| Toiviainen, 1995 | jazz improvisation | recurrent model |
| Nishijimi & Watanabe, 1993 | jazz improvisation | feedforward model |
| Franklin, 2001 | jazz improvisation | recurrent model |
| Hild et al., 1992 (HARMONET) | four-part harmonization | three-layered architecture (two ANNs and a constraint system) |
| Feulner & Hörnel, 1994 (MELONET) | four-part harmonization | uses HARMONET and another ANN for melodic variations |
| Goldman et al., 1996 (NETNEG) | species counterpoint | ANN for basic melody, an ensemble of agents refine the melody |
| Verbeurgt et al., 2004 | melody | two stages: a Markovian model and an ANN |
| Adiloglu & Alpaslan, 2007 | species counterpoint | feedforward model |
| Browne & Fox, 2009 | melody | simulated annealing with an ANN to measure musical tension |
| Coca et al., 2011 | melody | recurrent model, uses chaotic non-linear systems to introduce variation |

Table 13: References for Section 3.4, in order of appearance.

instead of absolute pitches (as Todd's work) in the mapping, for composing music in Bach's style.

As a machine learning paradigm, ANNs can be used in many different ways, so Todd's approach is not the only possible; indeed, other early papers provide different examples. For example, Mozer (1991) developed a recurrent ANN with a training program devised to capture both local and global patterns in the set of training examples. The model also featured in the output mapping a sophisticated multidimensional space for pitch representation, to capture a formal psychological notion of similarity between different pitches. In this way, similar output signals are mapped to similar pitches, in order to facilitate the learning phase and improve the composition phase. Lewis (1991) proposed another ANN framework: *creation by refinement*, in which a feedforward ANN was trained with a set of patterns ranging from random to very good music, associating each pattern with a (possibly) multidimensional musicality score. In this way, the training phase generated a mapping function from patterns to musicality scores. Then, to create new compositions, the mapping was inverted: starting from a purely random pattern, a gradient-descent algorithm used the ANN as a *critique*, reshaping the random pattern to maximize the musicality score in the hope of finally producing a pleasant composition. Unfortunately, this paradigm had a prohibitive computational cost, so it was tested only with fairly simple and short compositions.

Most of the early examples described above were experiments at composing more or less full-fledged monophonic compositions. However, ANNs were also used to automate other tasks in music composition, as harmonization of pre-existing melodies. Shibata (1991) implemented an early example: a feedforward ANN that represented chords using their component tones, trained for harmonizing simple MIDI music, whose performance was measured by human listeners. A more sophisticated ANN used for harmonization, the *effective Boltzmann machine* (EBM), also provided a measure of the quality of the output relative to the training set (Bellgard & Tsang, 1992). Melo (1998) also harmonized classical music, but with a notable twist: in order to model the tension[25] in the music to be harmonized, he measured the tension curve reported by several human subjects while listening to the music, and then used an averaged tension curve to train an ANN, such that the chord progressions generated by the ANN matched the tension level suggested by the curve. As it can be seen, harmonization was a popular test case, but other problems were also tried. For example, Toiviainen (1995) used ANNs to generated jazz improvisations based on a set of training examples. The ANNs were able to create new jazz melodic patterns based on the training set. In a similar way, Nishijimi and Watanabe (1993) trained a set of feedforward ANNs to produce jazz improvisations in a jam session, by modeling several music features of jazz and using examples of modeled jazz improvisations to train the ANNs. Franklin (2001) used a recurrent ANNs to improvise jazz (*trade four solos* with a jazz performer), trained in two phases: a first phase training an ANN with a set of pre-specified examples, and a second phase where the ANN is reconfigured and trained by reinforcement learning, where the reinforcement values are obtained by applying a set of heuristic rules.

Some researchers came to use hybrid systems, combining ANNs with other methods. One of the first examples was HARMONET (Hild et al., 1992): a model designed to solve a more complex task: four-part choral harmonization in Bach's style. HARMONET had a three-layered architecture: the first component was a feedforward ANN with a sophisticated encoding of musical information (optimized for harmonization functions instead of individual pitches), which was used to extract harmonization information. The output was fed to the second component, a rule-based constraint satisfaction algorithm to generate the chords, and the final component was another ANN designed to add quaver ornaments to the previously generated chords. As an evolution of HARMONET, MELONET (Feulner & Hörnel, 1994; improved by Hörnel & Degenhardt, 1997) not only harmonized chorales, but also generated melodic variations for their voices, using HARMONET as a first processing stage for the harmonization, then used another neural network to generate the melodic variations.

NETNEG (Goldman et al., 1996) was another hybrid system that used an ANN trained with sixteenth century classical music compositions. The ANN generated a basic melody by segments. After each segment was created, an ensemble of agents generated a polyphonic elaboration of the segment. The agents had rule-based systems crafted from music theoret-ical considerations, and coordinated to maintain a coherent global output. ANNs can also be combined with probabilistic methods: in the work of Verbeurgt et al. (2004), a set of training sequences was decomposed into musical motifs, encoded in relative pitch. Then, a Markov chain was constructed, whose states were the motifs. New compositions were generated by the Markov chain, but to assign the absolute pitches to the motifs in the resulting composition, they trained an ANN. Adiloglu and Alpaslan (2007) used feedforward ANNs to generate two-voice counterpoint, applying notions of music theory to the representation of musical information in the networks. In Browne and Fox's (2009) system, *simulated annealing* was used to arrange small fragments (motifs) of classical music, trying to get a profile of musical tension (the same metric as in Melo, 1998) similar to the profile of a pre-specified composition, measured using an ANN specialized in music perception. Finally, another hybrid system was implemented by Coca et al. (2011), which used ANNs trained with pre-existing compositions together with pseudo-random musical input generated from a chaotic system, in order to generate more complex compositions in the synthesis phase.

---

25. An important property of music, rather difficult to define. At each point in time, the tension is related to the interplay between structure and uncertainty perceived by the listener in the flow of the music. Informally, it can be defined as the *unfinishedness* of the music if it were stopped at that point.

### 3.4.1 ANNs WITH EVOLUTIONARY ALGORITHMS

Among hybrid systems, those combining ANNs with evolutionary algorithms quickly became the most popular. Usually, an ANN was trained to act as the fitness function of an evolutionary algorithm. This is the case of the earliest example of these hybrid systems, NEUROGEN (Gibson & Byrne, 1991). Its fitness function was the composed result of two ANNs, one for judging the intervals between pitches and the other for the overall structure. A genetic algorithm with a rather rigid, classical binary representation was used, severely limiting the applicability of the whole implementation. However, there are also "inverted" frameworks where the evolving individuals are ANNs. For example, Hörnel and Ragg (1996) evolved HARMONET networks, but the fitness was the network performance in training and harmonization. In another example (Chen & Miikkulainen, 2001), recurrent three-layered ANNs evolved to compose music, and the fitness was computed from a set of rules from music theory.

Given the modular nature of evolutionary algorithms and the perceived complexity of ANNs, it is not uncommon that the evolutionary framework is laid down in a first research work, and only in subsequent developments ANNs are used to replace the original fitness function.[26] For example, Spector and Alpern (1994) developed a *genetic programming* (GP) framework for jazz improvisations: the individuals were programs composed by collections of transformations that produced improvisations upon being fed previously existing jazz melodies, and the fitness function aggregated several simple principles from jazz music theory. The following year (Spector & Alpern, 1995), they updated their model to train an ANN to be used as their fitness function. However, this scheme does not always fare well, specially if the initial framework uses interactive fitness. This is the case of GenJam (Biles, 1994), an evolutionary algorithm for generating jazz melodies with an interactive fitness function. Later on (Biles et al., 1996), as the interactive fitness evaluation represented a severe fitness bottleneck, ANNs were tried to partially offload evaluation from human users, with no success, as the ANNs failed to satisfactorily generalize the evaluations from their training sets. In the case of the GP-Music System (Johanson & Poli, 1998), which used GP with procedural representations of short melodies, the trained ANNs were not a failure, but decidedly below par with respect to the performance of the algorithm with interactive fitness. Klinger and Rudolph (2006) compared the performance of feedforward ANNs with learned decision trees, finding that the latter performed better and their ratings were easier to understand. In spite of these examples, more successful instances do exist: Manaris et al. (2007) extracted several statistical metrics from music compositions and trained an ANN to recognize compositions whose metrics' distributions featured Zipf's law, then used it as the fitness function in a GP framework whose individuals were procedural representations of polyphonic compositions. The results were validated as aesthetically pleasing by human testers.

---

26. This approach is risky, though, because evolutionary algorithms tend to find and exploit unexpected and undesired quirks in any fitness evaluation function; more so if the evaluator is an ANN.

| Reference | Composition task | Comments |
|---|---|---|
| Gibson & Byrne, 1991 | melody, rhythm | Two ANNs to define the fitness function |
| Hörnel & Ragg, 1996 | melody harmonization | evolves HARMONET, the fitness is the ANN's performance in training and harmonization |
| Chen & Miikkulainen, 2001 | melody | evolves recurrent networks, the fitness is computed from a set of rules |
| Spector & Alpern, 1994 | melody | genetic programming, ANNs as fitness functions |
| Biles et al., 1996 | jazz improvisation | ANNs as fitness functions |
| Johanson & Poli, 1998 (GP-Music) | melody | genetic programming, ANNs as fitness functions |
| Klinger & Rudolph, 2006 | melody | ANNs and decision trees as fitness functions |
| Manaris et al., 2007 | melody | genetic programming, ANNs as fitness functions |
| Burton, 1998 | drum rhythms | Adaptive Resonance Theory (unsupervised learning) as fitness function |
| Phon-Amnuaisuk et al., 2007 | melody | genetic programming, self-organinzing map (unsupervised learning) as fitness function |

Table 14: References for Section 3.4.1, in order of appearance.

All the research described up to this point uses ANNs with supervised learning. However, methods using unsupervised learning also exist. Burton (1998) proposed a genetic algorithm with a classical binary representation for generating multi-voice percussion rhythms, whose fitness function presented an unconventional feature mix. It used *Adaptive Resonance Theory* (ART), an ANN with unsupervised learning, initially trained to perform automatic clustering of a set of drum rhythms. Then, during the execution of the genetic algorithm, the unsupervised learning continued. The fitness was a measure of how near was the individual to some cluster. If the individual represented a brand new rhythm so different as to add a new cluster, the rhythm was presented to the user to decide if it was musically acceptable. In this way, Burton tried to get the best of interactive and automatic fitness evaluation. A different example with unsupervised learning was presented by Phon-Amnuaisuk et al. (2007), able to generate variations over a pre-specified composition. It used GP whose individuals were procedural representations of melodies, while the fitness was the similarity to the pre-specified composition, as measured by a *self-organizing map* (SOM) previously trained with musical elements of the pre-specified composition.

| Reference | Composition task | Comments |
|---|---|---|
| Baggi, 1991 | jazz improvisation | *ad hoc* connectionist expert system |
| Laine, 2000 | simple rhytms | uses central pattern generators |
| Dorin, 2000 | poly-rhythmic musical patterns | uses Boolean networks |
| Hoover et al., 2012 | acompaniment | uses CPPNs |

Table 15: References for Section 3.4.2, in order of appearance.

### 3.4.2 Related Methods

Finally, this subsection presents methods that may be considered roughly similar to ANNs, or more generally, *connectionist*. For example, *Neurswing* (Baggi, 1991) may be described as an *ad hoc* expert system for jazz improvisation crafted in a connectionist framework. Surprisingly, many research papers cite *Neurswing* as an ANN framework, despite Baggi's disclaimer: "[Neurswing], though only vaguely resembling a neural net or a connectionist system, [. . . ]" (Baggi, 1991). Laine (2000) used very simple ANNs to implement *Central Pattern Generators*, whose output patterns where then interpreted as more or less simple rhythms (*motion patterns* in his terminology).

Boolean networks are another connectionist paradigm in which each node has a binary state and the edges between nodes are directed; each node's state changes in discrete steps according to a (usually randomly chosen) Boolean function whose inputs are the states of the nodes with connections to that node. They may be considered as generalizations of cellular automata, and depending on the wiring and the distribution of Boolean functions, their states can change in very complex patterns, with potentially complex responses to external forcing. Because of these properties, Dorin (2000) used Boolean networks to generate complex poly-rhythmic musical patterns, modulable in real time by the user.

Hoover et al. (2012) proposed another connectionist approach: compositional pattern producing networks (CPPNs). These are feedforward networks where each "neuron" may use a different, arbitrary function, instead of the classical sigmoid function. They are usually designed by interactive evolutionary methods, and can be used to generate or modulate highly complex patterns. In the cited paper, they were fed a pre-existing simple composition as input, in order to generate an accompaniment for it.

## 3.5 Evolutionary and Other Population-Based Methods

Most evolutionary algorithms (EAs) approximately follow a common pattern: a changing set of candidate solutions (a population of individuals) undergoes a repeated cycle of evaluation, selection and reproduction with variation. The first step is to generate the candidate solutions of the initial set, either from user-specified examples or in a more or less random way. Each candidate is then evaluated using a *fitness* function, a heuristic rule to measure its quality. The next phase is selection: a new set of candidate solutions is generated by copying candidate solutions from the old one; each candidate solution is copied a number of times probabilistically proportional to its fitness. This step decreases the diversity of the population, which is restored by applying (to a fraction of the candidate solutions) some operators designed to increase the variation (for example, mutation or recombination operators). These steps are applied iteratively; as a result, best and mean fitness gradually tend to increase.

While this algorithmic pattern is common to all EAs, there exist many different algorithms using different sets of selection rules, variation operators and solution encoding. In an EA, the encoded form of a candidate solution is the genotype, while the phenotype is the translation of that coded form into a solution. Holland's original formulation of *genetic algorithms* is strongly associated with a plain and direct encoding of genotypes as binary strings, but this is not the case in most papers using EAs. Because of this, the term *evolutionary* is preferred over *genetic* in this paper. Another popular variant, Koza's *genetic programming*, represents genotypes as tree structures, often encoding expressions of a programming language, while the phenotype is the result of evaluating these expressions.

Since EAs are particularly prone to be hybridized with other methods, they are also reviewed in the other sections: with grammars in Section 3.1.2, with ANNs in Section 3.4.1, with Markov chains in Section 3.3[†], with rule-based systems in Section 3.2.2, and with cellular automata in Section 3.6.1[‡]. In general, the papers cited in these sections will not be discussed here. We also recommend some literature to learn more about evolutionary computer music: Burton and Vladimirova (1999) wrote a survey with long, thorough descriptions of the referenced papers, while the survey of Santos et al. (2000) is more similar to ours in style, packed with brief descriptions. Miranda and Biles's (2007) book is more recent, but also contains work on other optimization methods (as swarm optimization) and algorithmic composition techniques (as cellular automata).

### 3.5.1 Evolution with Automatic Fitness Functions

The difficulty to define automatic fitness functions has been a constant issue, frequently limiting the application of evolutionary methods to well-defined and restricted problems in composition. Horner and Goldberg (1991) provided one of the first examples: they implemented an EA for *thematic bridging*, a composition technique consisting of defining a sequence (with a pre-specified preferred length) of small musical patterns such that the first and the last ones are pre-specified, and each pattern is the result of applying a simple transformation to the previous one. Naturally, the individuals in the EA were defined as lists of operations applied to the initial pattern to generate the sequence of patterns. The fitness measured how close the final pattern (generated from the operations) was to the pre-specified final pattern, plus the difference between actual and preferred lengths of the sequence. Essentially the same work (with differences in the underlying representation and operation set) was reported by Ricanek et al. (1993).

A common way to implement a fitness function is as a weighted sum of features of the composition (although tuning the weights to optimize the EA can prove difficult except for toy problems). For example, Marques et al. (2000) composed short polyphonic melodies using a very direct representation for the genotypes and also a simple fitness function, with a rather simple, *ad hoc* evaluation of harmony and melodic value. The results were reportedly acceptable. Johnson et al. (2004) also composed short melodies using an EA with a fitness function that was a weighted sum of a series of very basic, local features of the melody. Papadopoulos and Wiggins (1998) implemented a system that, given a chord progression, evolved jazz melodies by relative pitch encoding, using as fitness function a weighted sum of eight evaluations of characteristics of the melody, ranging from very simple heuristics about the speed and the position of the notes to user-specified contour and similarity to user-specified music fragments. A similar approach was implemented by Harris (2008), with modest (if promising) results. While Towsey et al. (2001) did not actually implement an EA, they discussed how to build a fitness function for melodic extension (given a composition, extend it for a few bars): they proposed to extract 21 statistical characteristics from a corpus of pre-specified compositions, defining the fitness function as a weighted sum of the distance between the individual and the mean of the characteristic. In a similar vein, Birchfield (2003) implemented a fitness function as a giant weighted sum of many features in a hierarchical EA, with multiple population levels, each individual in a population being composed of individuals from lower populations (similar to the model described in Biles, 1994; see Section 3.5.2). He used the output of the EA as material to arrange a long composition for ten instruments. Garay Acevedo (2004) implemented a simple EA to compose first species counterpoint, but the features weighted in the fitness function were too simplistic, leading to modest results. Lozano et al. (2009) generated chord sequences to harmonize a pre-specified melody in two steps: first a simple EA generated a set of possible solutions according to simple local considerations (appropriateness of each chord for the corresponding part of the melody) between the chords and the notes of the melody, and then a variable neighborhood search was used to establish a chord progression according to global considerations.

---

†. Only a few examples of evolutionary algorithms were described in this section, not warranting a dedicated subsection for them.

‡. Same case as in the previous note.

| Reference | Composition task | Comments |
|---|---|---|
| Horner & Goldberg, 1991 | thematic bridging | fitness: distance to original melodies |
| Ricanek et al., 1993 | thematic bridging | fitness: distance to original melodies |
| Marques et al., 2000 | polyphony | fitness: combination of features |
| Johnson et al., 2004 | melody | fitness: combination of features |
| Papadopoulos & Wiggins, 1998 | jazz improvisation | fitness: combination of features |
| Harris, 2008 (JazzGen) | jazz improvisator | fitness: combination of features |
| Towsey et al., 2001 | melodic extension | fitness: combination of features (only fitness, no evolutionary algorithm) |
| Birchfield, 2003 | melody, rhythm | fitness: combination of features |
| Garay Acevedo, 2004 | species counterpoint | fitness: combination of features |
| Lozano et al., 2009 | melody harmonization | fitness: combination of features |
| De Prisco et al., 2010 | unfigured bass | multi-objective optimization |
| Freitas & Guimarães, 2011 | melody harmonization | multi-objective optimization |
| Gartland-Jones, 2002 | generate variations on two melodies | fitness: distance to a melody |
| Alfonseca et al., 2005 | melody | fitness: distance to corpus of melodies |
| Özcan & Erçal, 2008 (AMUSE) | generate variations on a melody | fitness: combination of features |
| Wolkowicz et al., 2009 | melody | fitness: distance to corpus of melodies |
| Laine & Kuuskankare, 1994 | melody | fitness: distance to a melody. Genetic programming |
| Spector & Alpern, 1994 | jazz improvisation | fitness: combination of features. Genetic programming |
| Dahlstedt, 2007 | contemporary classical music | fitness: combination of features. Genetic programming |
| Espí et al., 2007 | melody | fitness: combination of features extracted fropm a corpus of melodies. Genetic programming |
| Jensen, 2011 | melody | fitness: distance to a corpus of melodies. Genetic programming |
| Díaz-Jerez, 2011; Sánchez-Quintana et al., 2013 | contemporary classical music, other genres | sophisticated indirect encoding |

Table 16: References for Section 3.5.1, in order of appearance.

The alternative to implementing the fitness function as a weighted sum of musical features is to use *multi-objective evolutionary algorithms* (MOEAs). However, these have very rarely been used, most probably because they are harder to implement, both conceptually and in practice. MOEAs have been used by De Prisco et al. (2010) to harmonize the unfigured bass[27] and by Freitas and Guimarães (2011) for harmonization.

Another approach consists of measuring the fitness as the distance to a target composition or corpus of compositions. For example, Gartland-Jones (2002) implemented an EA to compose hybrids between two pre-specified compositions (a goal similar to Hamanaka et al.'s, 2008, cited in Section 3.1) by using one to seed the initial population, and the distance to the other (sum of difference in pitch for each note) as the fitness function. Alfonseca et al. (2005) used a more sophisticated evaluation: the fitness of each composition in the population was the sum of distances to a corpus of pre-specified target compositions. The metric was the normalized compression distance, a measure of how different two symbol strings are, based on their compressed lengths (both concatenated and separated). Özcan and Erçal (2008) used a simple genetic representation and a fitness function based on a weighted sum of a long list of simple musical characteristics, reportedly being able to generate improvisations over a pre-specified melody with a given harmonic context, but they did not specify how the evolved melodies were related to the pre-specified music. In the EA implemented by Wolkowicz et al. (2009), individuals were encoded using relative pitches, and a sophisticated statistical analysis of $n$-gram sequences in a pre-specified corpus of compositions was used to implement the fitness function.

Genetic programming has also been used with fitness functions based on comparisons with pre-specified music. In the work of Laine and Kuuskankare (1994), the individuals were discrete functions of time whose output (phenotype) was interpreted as a sequence of pitches. The fitness was simply the sum of differences in pitches between a pre-specified target composition and the phenotype of each individual, in a similar way to Gartland-Jones (2002). However, it is frequent to find fitness functions based on the analysis of characteristics of the compositions evolving in the algorithm, often comparing them against characteristics of a pre-specified training set of compositions. Spector and Alpern (1994) used genetic programming for jazz improvisation, "trading fours": the individuals were functions that took as input a "four" and produced another one by applying a series of transformations to it. The fitness was determined by applying a series of score functions that measured the rhythm, tonality and other features of the produced "four", comparing them against a database of high-quality examples from renowned artists.

---

27. For a description of the unfigured bass, see the discussion of Rothgeb's (1968) work in Section 3.2.

Also using genetic programming, Dahlstedt (2007) composed relatively short contemporary classical music pieces, with a simple fitness function: a set of target values were assigned for several statistics of the compositions (as note density or pitch standard deviation, among others), and the fitness was a weighted sum of the differences between the target values and the values for each individual. The individuals were trees whose nodes represented notes and different operations over musical sequences (using a developmental process from genotype to phenotype). Reportedly, the generated pieces were of acceptable quality, because the genotype model was especially well suited for contemporary classical music. Espí et al. (2007) used a simple tree representation for compositions (but without indirect encoding), defining the fitness function as a weighted sum of sophisticated statistical models for melody description (measuring the distance to the values of a set of pre-specified compositions) and several relatively simple characteristics of the composition. Jensen (2011) also used a simple tree representation for compositions; the fitness was calculated measuring frequency distributions of simple events in the compositions, rating them according to Zipf's law and similarity to pre-specified compositions.

One of the latest and most successful results in evolutionary computer music follows an evo-devo strategy. Iamus is a computer cluster which hybridizes bioinspired techniques: compositions evolve in an environment ruled by formal constraints and aesthetic principles (Díaz-Jerez, 2011). But compositions also develop from genomic encodings in a way that resembles embryological development (hence the evo-devo), providing high structural complexity at a relatively low computational cost. Each composition is the result of an evolutionary process where only the instruments involved and a preferred duration have been specified, and are included in the fitness function. Iamus can write professional scores of contemporary classical music, and it has published its debut album in September 2012 (Ball, 2012; Coghlan, 2012), with ten works interpreted by first-class musicians (including the LSO for the orchestra piece). Melomics, the technology behind this avant-garde computer-composer, is also mastering other genres and transferring the result to industry (Sánchez-Quintana et al., 2013). After compiling a myriad of musical fragments of most essential styles in a browsable, web-based repository (Stieler, 2012). For the first time, Melomics is offering music as a real commodity (priced by size of its MIDI representation), where ownership over a piece is directly transferred to the buyer.

### 3.5.2 Musical IGAs

From the previous exposition, it is apparent that designing an objective and convenient fitness function for evaluating music compositions is a very difficult problem.[28] If the music is to be evaluated in terms of subjective aesthetic quality, it may become impractical or directly impossible to define a formal fitness function. Because of these inconveniences, many researchers have resorted to implement the fitness function with human evaluators. A common term for describing this class of EAs is *musical IGA* (interactive genetic algorithm[29], MIGA for short). As MIGAs represent a substantial percentage of the total body of work on EAs for algorithmic composition, this subsection is devoted to them.

---

28. To the point that artists sometimes find "unconventional" ways around this problem: Waschka (1999) argued to have solved the problem by assigning a *purely random* fitness value to the individuals.

| Reference | Composition task | Comments |
|---|---|---|
| Hartmann, 1990 | melody | inspired by Dawkins' biomorphs |
| Nelson, 1993 | melody | inspired by Dawkins' biomorphs |
| Horowitz, 1994 | rhythms | inspired by Dawkins' biomorphs |
| Pazos et al., 1999 | rhythms | binary genotype |
| Degazio, 1996 | melody | binary genotype; graphical representation |
| Biles, 1994 (GenJam) | jazz improvisation | two hierarchically structured populations (measures and jazz phrases) |
| Tokui & Iba, 2000 | rhythms | two hierarchically structured populations (short and long rhythmic patterns). Genetic programming |
| Jacob, 1995 | melody | the user trains "critics" that act as fitness functions |
| Schmidl, 2008 | melody | the user trains "critics" that act as fitness functions |
| Putnam, 1994 | melody | genetic programming |
| Ando & Iba, 2007 | melody | genetic programming |
| MacCallum et al., 2012 (DarwinTunes) | melody | genetic programming |
| Kaliakatsos-Papakostas et al., 2012 | melody, 8-bit sound synthesis | genetic programming |
| Hoover et al., 2012 | acompaniment | uses CPPNs |
| McDermott & O'Reilly, 2011 | interactive generative music | similar to CPPNs |
| Ralley, 1995 | melody | minimizes user input by clustering candidates |
| Unehara & Onisawa, 2001 | melody | minimizes user input with elitism |
| Díaz-Jerez, 2011 | contemporary classical music | minimizes user fatigue producing small, good compositions |
| Beyls, 2003 | melody | uses cellular automata; graphical representation |
| Moroni et al., 2000 (Vox Populi) | melody | complex graphical representation |
| Ventrella, 2008 | melody | the whole population comprises the melody |
| Marques et al., 2010 | melody | minimizes evolutionary iterations |

Table 17: References for Section 3.5.2, in order of appearance.

The first MIGAs were implemented by composers intrigued by the concept of evolutionary computing, resulting in more or less peculiar architectures from the perspective of common practice in evolutionary computing, but also by computer scientists exploring the field. Hartmann (1990), inspired by Dawkins' biomorphs, presented one of the first applications of evolutionary computing to composition, a MIGA that he unfortunately described in a notoriously laconic and obscure language, resulting in a very low citation rate for his work. Also inspired by the biomorphs, Nelson (1993) described a toy MIGA for evolving short rhythms over a fixed melodic structure, with simple binary genotypes (each bit simply denoted the presence or absence of sound). More formal models similar in scope to Nelson's were designed by Horowitz (1994) and Pazos et al. (1999). They implemented rhythm generators for multiple instruments, each one with its own independent rhythm pattern encoded in the genotype (Horowitz's genotypes were parametric, while Pazos et al. used more direct binary encodings). Degazio (1996) implemented a system in which the genotype was a set of parameters (in later iterations, a mini-language to describe the parameters) to instruct his CAAC software to generate melodies.

---

29. Most research tagged as IGA does not use the binary genotypes commonly associated with the term *genetic algorithm*, but the term is very common.

The best known MIGA may be GenJam, a system for generating jazz solos, developed over several years. In its first incarnation (Biles, 1994), it was formulated as a MIGA with two hierarchically structured populations: one of measures, and other of jazz phrases, constructed as sequences of measures. Given a chord progression and several other parameters, jazz solos emerged by concatenating selected phrases during the evolutionary process, and the fitness was integrated over time by accumulating fixed increments and decrements from simple good/bad indications from the evaluator. Further iterations of the system included the already discussed use of ANNs as fitness functions (Biles et al., 1996) and the possibility to *trade fours* with a human performer, by dynamically introducing into the population the music performed by the human (Biles, 1998). Tokui and Iba (2000) used a similar solution for creating rhythms with multiple instruments: a population of short sequences specified as list of notes, and another population of tree structures representing functions in a simple macro language that used the short sequences as building blocks. Another example of hierarchical structuring of the MIGA is Jacob's (1995) system for general-purpose composition, with three inter-dependent evolutionary processes: one involving the human user to train *ears* to evaluate short musical sequences, another one to compose musical phrases using the *ears* (filters) as fitness functions, and another also involving the human user to train an *arranger* that reorders the resulting phrases into the final output of the system. Schmidl (2008) implemented a similar system, but without a high-level *arranger* module, and with *ears* automatically trained from a set of examples, in order to minimize user interaction and enable real-time composition.

Genetic programming with interactive evaluation has been used several times. Putnam (1994) implemented an early example: each individual coded for a set of functions that generated a melody as the result of an iterated function system. Tokui and Iba's (2000) example has been already cited. Ando and Iba (2007) implemented a fully interactive system (not only the selection, but also the reproduction and mutation were user-guided), where the genotype model was similar to Dahlstedt's (2007). MacCallum et al. (2012) used trees to encode Perl expressions that generated polyphonic short loops, but concentrated on analyzing the interactive evolution from the point of view of theoretical biology. Kaliakatsos-Papakostas et al. (2012) used a rather different approach to generate 8-bit melodies: each individual was a function composed of bitwise operators that generated a waveform by iterating the function. In fact, that work might be described as sound synthesis over large time scales, rather than music composition.

Some MIGAs with graph-based genetic representations also exist, as the implementation based in CPPNs (Hoover et al., 2012) that was already cited in Section 3.4.2. McDermott and O'Reilly (2011) used a similar paradigm: genotypes were sequences of integers that indirectly encoded graphs, whose nodes represented functions, and their connections were compositions of functions. The output nodes generated musical output in one or more voices, which were modulated by user inputs.

A problem common to all MIGAs is user fatigue: candidate solution evaluation is a comparatively slow and monotone task that rapidly leads to user fatigue. Even with small population sizes and small numbers of generations, it remains a significant problem that has been solved by many researchers in different ways. For example, Ralley (1995) used a binary representation with relative pitch encoding for the genotypes, and classified the population with a clustering algorithm, deriving similarity metrics from rudimentary spectral analysis of the scores. The user was simply required to evaluate the closest composition to the centroid of each cluster. A more exotic solution was employed by previously cited Tokui and Iba (2000): training a neural network to filter out candidates with low fitness, thus presenting to the user individuals of acceptable quality. Unehara and Onisawa (2001) presented just 10% of candidate melodies to the human user, and then parts of the genomes of the best rated ones were dispersed in the population by "horizontal gene transfer". McDermott and O'Reilly (2011) also limited the number of candidates exposed to user rating, filtering out the worst ones with heuristic functions. Another option to minimize user fatigue is to produce small compositions that already are reasonably good. The Melomics system (see last paragraph in Section 3.5.1) can be used in this way (Díaz-Jerez, 2011).

Other common ways to manage the problem include low population sizes and/or hierarchical structuring of the algorithm (Biles, 1994), or providing statistical information and/or rendering graphical representations of the compositions in order to make possible their evaluation without actually listening to them (Degazio, 1996). Graphical representations are also particularly useful if using generative methods as L-systems or cellular automata (Beyls, 2003). Putnam (1994) used a web interface to reach out to more volunteers. Moroni et al. (2000) tried to solve the problem using sophisticated GUI abstractions, with complex non-linear mappings between the graphic controls and the parameters of the fitness function and other aspects of the evolutionary process, to produce a highly modulable system for real-time interactive composition of melodies. To mitigate user fatigue, Ventrella (2008) presented a population of short melodies as a continuous stream of sound; fitness was obtained through a binary signal set by the user. Marques et al. (2010) limited user fatigue in the generation of short, simple melodies by severely limiting the number of generations of the algorithm, and generating a reasonably good starting population by drawing the notes using Zipf's law.

### 3.5.3 Other Population-Based Methods

Finally, this subsection presents other methods that are also population-based. For example, the metaheuristic method *Harmony Search* is inspired in the improvisation process of musicians, though in practice can be framed as an evolutionary method with a specific way to structure candidate solutions and perform selection, crossover and mutation operations. Geem and Choi (2007) used this method to harmonize Gregorian chants (i.e., to write or-ganum lines for the chants). Other methods are also population-based but not properly evolutionary. An example is the use of *Ant Colony Optimization* (ACO) to solve constraint harmonization problems (Geis & Middendorf, 2008), already mentioned in Section 3.2.2. In ACO, the candidate solutions are represented as paths in a graph, and a population of agents (ants) traverse the graph, cooperating to find the optimal path.

| Reference | Composition task | Comments |
|---|---|---|
| Geem & Choi, 2007 | harmonize Gregorian chants | harmony search |
| Geis & Middendorf, 2008 | four-part harmonization | multi-objective Ant Colony Optimization |
| Tominaga & Setomoto, 2008 | polyphony, counterpoint | artificial chemistry |
| Werner & Todd, 1997 | melody | co-evolutionary algorithm, Markov chains used as evolvable fitness functions |
| Bown & Wiggins, 2005 | melody | individuals are Markov chains that compose and evaluate music |
| Miranda, 2002 | melody | individuals agree on a common set of "intonation patterns" |
| Miranda et al., 2003, sect. IV | melody | individuals are grammars that compose music |
| McCormack, 2003b | interactive soundscape | individuals indirectly compete for users' attention |
| Dahlstedt & Nordahl, 2001 | soundscape | music emerges from collective interactions |
| Beyls, 2007 | soundscape | music emerges from collective interactions |
| Blackwell & Bentley, 2002 | soundscape | music emerges from collective interactions |
| Eldridge & Dorin, 2009 | soundscape, sound synthesis | music emerges from collective interactions, individuals exist in the frequency domain |
| Bown & McCormack, 2010 | interactive soundscape, sound synthesis | music emerges from collective interactions, individuals exist in the frequency domain |

Table 18: References for Section 3.5.3, in order of appearance.

As a more exotic example, an *Artificial Chemistry* is a generative system consisting of a multiset of strings of symbols. These strings (analogues of molecules) can react according to a pre-specified set of rules (analogues of chemical reactions), generating new strings from the existing ones. Tominaga and Setomoto (2008) used this method, encoding polyphonic compositions in the strings and musical rules for counterpoint in the reaction rules of the artificial chemistry: starting from a set of simple strings, the system generated progressively more complex ones, though the aesthetical value of the resulting compositions varied widely.

A more popular method based on populations of individuals is the *Artificial Ecosystem*. In an artificial ecosystem, compositions emerge from the interaction between individuals in a simulation with evolutionary and/or cultural interactions, taking inspiration in the evolutionary origins of music in humans (Wallin & Merker, 2001). Frequently, the complexity of the simulations is severely limited by the available computational power, and in some cases the goal is not music composition *per se*, but the study of evolutionary dynamics, the emergence of shared cultural traits and *avant-garde* artistic experimentation.

An early example by Werner and Todd (1997) investigated sexual evolutionary dynamics in a population of males (small compositions) and females (Markov chains initially generated from a corpus of songs) that evaluated how much the males deviated from their expectations. However, most studies use just one kind of agent, as in the work of Bown and Wiggins (2005), whose agents used Markov chains for both compose and analyze music. Miranda (2002) and Miranda, Kirby, and Todd (2003, sect. IV) implemented models with a similar goal: to study the emergence of common structures (shared cultural knowledge). In the first case a population of agents strove to imitate each other's intonation patterns (short sequences of pitches); in the second case agents learned to compose music by inferring musical grammars from other agents' songs. Bosma (2005) extended Miranda's (2002) model, using neural networks in the agents to learn and compose music, although with tiny population sizes. As part of an art installation, McCormack (2003b) proposed a virtual ecosystem with evolving agents able to compose music using a rule-based system, competing for resources that were indirectly determined by the interest of human observers.

An alternative is that the music is not composed by the agents, but emerges as an epiphenomenon of the whole ecosystem. The models of Dahlstedt and Nordahl (2001) and Beyls (2007) used simple organisms in a two-dimensional space whose collective behavior was mapped into complex compositions, while Blackwell and Bentley's (2002) model was similar but three-dimensional, and the dynamics of the agents were inspired in swarm and flocking simulations. While all these examples use either homogeneous or spatially structured ecosystems, a recent trend is the use of sound as an environment in itself. In Eldridge and Dorin's (2009) model, the agents dwelt in the one-dimensional space of the Fourier transform of a sample of ambient sound, feeding off and moving the energy across frequencies. Bown and McCormack (2010) implemented a similar model, in which the agents were neural networks that generated sound and competed for room in the space of frequencies of ambient sound.

### 3.6 Self-Similarity and Cellular Automata

In the late 1970s, two notable results about music were reported by Voss and Clarke (1978). The first was that, for music of many different styles, the spectral density of the audio signal was (approximately) inversely proportional to its frequency; in other words, it approximately follows a $1/f$ distribution. This is not so surprising: many different data series follow this property, from meteorological data to stock market prices; it is usually referred to as $1/f$ noise or *pink* noise. The second result was that random compositions seemed more musical and pleasing (for a wide range of evaluators, from unskilled people to professional musicians and composers) when the pitches were determined by a source of $1/f$ noise, rather than other common random processes as *white* (uncorrelated) noise or Brownian motion (random walks). Although the first result has been since challenged[30], the second one has been used by composers as a source of raw material. Bolognesi (1983) implemented an early example influenced by Voss and Clarke's results, but composers used data series with $1/f$ noise as raw material even *before* these results, early in the 1970s (Doornbusch, 2002).

---

30. The main criticism is that the data samples used by Voss and Clarke were hours long, merging in each sample many different compositions (and even non-musical sounds from a radio station). In view of their

| Reference | Composition task | Comments |
|---|---|---|
| Voss & Clarke, 1978 | melody | first reference to $1/f$ noise in music |
| Bolognesi, 1983 | melody | early deliberate use of $1/f$ noise in music |
| Doornbusch, 2002 | melody | reference to early non-deliberate use of $1/f$ noise in music |
| Gogins, 1991 | melody | iterated function systems |
| Pressing, 1988 | melody, sound synthesis | chaotic non-linear maps |
| Herman, 1993 | melody | chaotic non-linear dynamical systems |
| Langston, 1989 | melody | fractional Brownian motion |
| Díaz-Jerez, 2000 | melody | fractals and other self-similar systems |
| Bidlack, 1992 | melody | various fractal and chaotic systems |
| Leach & Fitch, 1995 (XComposer) | melody, rhythm | uses various fractal and chaotic systems |
| Hinojosa-Chapel, 2003 | melody | uses various fractal and chaotic systems to fill the |
| Coca et al., 2011 | melody | uses chaotic systems to add variation |

Table 19: References for Section 3.6, in order of appearance.

There remains the question of why $1/f$ noise produces more musical results that other random processes. The consensus in the research and artistic communities is *self-similarity*: the structure of $1/f$ noise is statistically similar across several orders of magnitude (Farrell et al., 2006). Self-similarity is a common feature in classical music compositions (Hsü & Hsü, 1991), and is also one of the defining features of *fractals* (in fact, $1/f$ noise also has fractal characteristics). Because of this, fractals have been extensively used as a source of inspiration and raw material for compositions and CAAC software. In general, self-similar musical patterns have multiple levels of structure, with pleasing regularities but also dotted with sudden changes. Because these characteristics can be also present in the output of chaotic systems (whose attractors are also fractal structures), these are also used to generate musical patterns. Commonly used techniques to generate self-similar musical patterns include chaotic systems such as iterated function systems (Gogins, 1991), non-linear maps (Pressing, 1988) and non-linear dynamical systems (Herman, 1993), but also fractional Brownian motion (Langston, 1989), cellular automata (discussed below) and L-systems (already discussed in Section 3.1.1). More exotic methods are also possible, as musical renderings of fractal images or number sequences with fractal characteristics (Díaz-Jerez, 2000). These methods are widely regarded as not suitable to produce melodies or compositions in their own right, but as a source of inspiration or raw material (Bidlack, 1992). Because of this, no extensive review will be provided here.[31] However, full-fledged algorithmic composition methods can use them as part of the creative process, as in Leach and Fitch's (1995) XComposer, where chaotic systems are used to fill the structures laid

---

critics, such as Nettheim (1992), these samples could not possibly be representative of single musical pieces (see also the discussion in Díaz-Jerez, 2000, pp. 136–138).

31. Good (if somewhat outdated) reviews can be found in the work of Jones (1989), Díaz-Jerez (2000) and Nierhaus (2009). The list of composers using fractals and chaotic systems for CAAC is so long that it is impractical to consistently describe all the existing relevant work.

down by a hierarchical model, or in Hinojosa-Chapel's (2003) paradigm for interactive systems, where they are also used as a source of musical material. They can also be used to add complexity to compositions generated by other means.[32]

### 3.6.1 Cellular Automata

A cellular automaton (CA) is a discrete (in time, space and state) dynamic system composed of very simple computational units (*cells*) usually arranged in an ordered n-dimensional (and potentially unbounded) grid (or any other regular tiling). Each cell can be in one of a finite number of states. In each discrete time step, each cell's state is deterministically updated, using a set of *transition rules* that take into account its own state and its neighbors' states. Although this definition can be generalized in multiple ways, it represents a good first approximation. Cellular automata are used in many disciplines across Science and the Humanities as dynamical models of complex spatial and temporal patterns emerging from the local interaction of many simple units; music composition is just one of these disciplines. Cellular automata can be used to generate fractal patterns and discrete versions of chaotic dynamical systems, but they also represent an alternative computational paradigm to realize algorithmic composition.[33] Unfortunately, just like fractals and chaotic systems, CA also tend to produce interesting but somewhat unmusical patterns that are used as inspiration or raw material rather than directly as music compositions. Although CA are argued to be better suited to sound synthesis than to algorithmic composition (Miranda, 2007), only the latter application will be reviewed here.

Xenakis was known to be deeply interested in the application of CA to music. In his orchestral composition *Horos*, released in 1986, he is widely regarded to have used a CA to configure the structure of the composition, though it was then heavily edited by hand (Hoffmann, 2002). Early, better documented explorations of CA for music composition include the implementations of Beyls (1989), Millen (1990), and Hunt et al. (1991), which mapped the patterns generated by user-defined CA to MIDI output. Beyls (1989) presented CA as a generative system for real-time composition of *avant-garde* music, exploring several ways to complexify the generated musical patterns (as changing the transition rules according to meta-rules), while Millen (1990) presented a minimalist CAAC system. Hunt et al. (1991) implemented another CAAC system designed to give the composer more control over the composition process. Echoing Beyls's (1989) early work, Ariza (2007) proposed to *bend* the transition rules in order to increase the space of parameters available to the composer for experimentation, by either randomly changing the state of some isolated cells or dynamically changing the transition rules from one generation to the next.

CAMUS (Miranda, 1993) is a more known CA system for algorithmic composition with an innovative design using two bidimensional CA: Conway's Game of Life (used to determine musical sequences) and Griffeath's Crystalline Growths (used to determine the instrumentation of the notes generated by the first CA). Each activated cell in the Game of Life was mapped to a sequence of three notes, whose instrument was selected according to the corresponding cell in the second CA (the Crystalline Growths system). Unfortunately, according to its own creator (Miranda, 2007), CAMUS did not produce very musical results: its out-put was more properly considered as raw material to be edited by hand. CAMUS was later generalized, using a Markov chain to determine the note durations and three-dimensional versions of the Game of Life and Crystalline Growths (McAlpine et al., 1999).

---

32. See, e.g., the description of the work by Coca et al. (2011) in Section 3.4.
33. For a more detailed survey, see e.g. the work of Burraston and Edmonds (2005).

| Reference | Composition task | Comments |
|---|---|---|
| Hoffmann, 2002 | structure | reference to early use of CA in music (Xenakis's *Horos*) |
| Beyls, 1989 | melody | early use of CA in music |
| Millen, 1990 | melody | early use of CA in music |
| Hunt et al., 1991 | melody | early use of CA in music |
| Ariza, 2007 | melody | dynamically changing CA rules |
| Miranda, 1993 (CAMUS) | melody, instrumentation | two CA: one for the melody, other for the instrumentation |
| McAlpine et al., 1999 (CAMUS 3D) | melody, rhythm, instrumentation | same as above, plus Markov chains to select rhythm |
| Bilotta & Pantano, 2001 | melody | explores several mappings from CA to music events |
| Dorin, 2002 (Liquiprism) | rhythmic patterns | several interacting CA |
| Ball, 2005, Miljkovic, 2007 | melody, rhythm | references to WolframTones |
| Phon-Amnuaisuk, 2010 | melody | uses ANNs to learn CA rules |
| Beyls, 2003 | melody | interactive evolutionary algorithm |
| Bilotta & Pantano, 2002 | melody | extends Bilotta and Pantano's (2001) work with an evolutionary algorithm |
| Lo, 2012 | melody | evolutionary algorithm, Markov chains used in the fitness function |

Table 20: References for Section 3.6.1, in order of appearance.

More recently, Bilotta and Pantano (2001) explored several different mappings to generate music from CA: *local codes* (mapping cells to pitches, the usual mapping in most papers), *global codes* (mapping the entropy of the whole pattern in each generation to musical events) and *mixed codes* (mapping groups of cells to musical events). Dorin (2002) used six bidimensional finite CA arranged in a cube (their edges connected), running at different speeds, to generate complex poly-rhythmic patterns.[34] Finally, WolframTones[35] (Ball, 2005) is a commercial application of CA to music composition, using a database of four billions of transition rules for one-dimensional CA (all possible transition rules taking into account five neighbors). WolframTones searches for rules that produce chaotic or complex patterns. These patterns are mapped to musical events, and the system is able to search for patterns whose musical mapping resembles one of a set of pre-defined musical styles (Miljkovic, 2007).

Although CA are commonly used to generate musical material in a "uncontrolled" way (i.e., the composer tunes the parameters of the CA by hand), it is possible to use other methods to design the CA (states, transition rules, etc.). For example, Phon-Amnuaisuk (2010) used artificial neural networks trained to *learn* the transition rules of a CA: given a melody, its piano-roll notation was interpreted as the temporal pattern of a CA, and the network was trained to learn the transition rules for that temporal pattern. Then, given other initial conditions, the network produced new compositions in piano-roll notation. Evolutionary algorithms are also used to design the parameters (transition rules, states, etc.) of CA. In some cases, previous work with hand-designed CA is adapted to use an evolutionary algorithm. This is the case of Beyls (2003), who used an interactive evolutionary algorithm to evolve the parameters of a CA, and Bilotta and Pantano (2002), who adapted their previously discussed work (Bilotta & Pantano, 2001) to use an evolutionary algorithm, although the fitness function was poorly described. Lo (2012) applied evolutionary algorithms to generate CA for algorithmic composition in a more comprehensive way, experimenting with various fitness functions based on extracting statistical models from a corpus of pre-existing compositions, including metrics based on Markov models and Zipf's law.

---

34. In Section 3.4.2, similar work (Dorin, 2000) with Boolean networks (a connectionist paradigm usually seen as a generalization of CA) was mentioned.

35. `http://tones.wolfram.com/`

## 4. Conclusions

In this survey, several hundreds of papers on algorithmic composition have been briefly reviewed. Obviously, none of them has been described in detail. Rather, this survey has been intended as a short reference guide for the various methods commonly used for algorithmic composition. As Pearce et al. (2002) noted, most papers on algorithmic composition do not adequately (a) specify the precise practical or theoretical aims of research; (b) use a methodology to achieve these aims; or (c) evaluate the results in a controlled, measurable and repeatable way. Researchers of algorithmic composition have very diverse backgrounds, and, in many cases, they do not present their work in a way that enables comparison with others. Because of these considerations, we have presented the literature in a "narrative" style, classifying the existing work in several broad categories, and providing brief descriptions of the papers in an approximately chronological order for each category.

### 4.1 About Creativity

Algorithmic composition automates (to varying degrees) the various tasks associated with music composition, such as the generation of melodies or rhythms, harmonization, counterpoint and orchestration. These tasks can be applied in two ways: (a) to generate music imitating a corpus of compositions or a specific style, and (b) to automate composition tasks to varying degrees, from designing mere tools for human composers, to generating compositions without human intervention:

> **Generating music imitating a corpus of compositions or a specific style.** Most instances of this kind of problem (including real-time improvisation systems that elaborate on input from human musicians) can be considered as solved: imitation problems have been tackled with many different methods, in many cases with reasonable success (such as Cope's EMI, 1992, or Pachet's Continuator, 2002). In fact, since the origins of computational algorithmic composition, there has been a bias in the research community towards imitation problems (Nierhaus, 2009). This may be attributed to the difficulty to merge the mindset of computer science (clear-cut definitions, precise algorithms, straight methodologies) with the mindset of artistic work (intuition, vagueness, cultural heritage and artistic influences). These two mindsets may be compared to the once common cultural divide in Artificial Intelligence between *neats* and *scruffies*. Unfortunately, while the neats reign supreme in Artificial Intelligence, they have yet to gain the upper hand in Artificial Creativity.

**Automating composition tasks to varying degrees.** In the case of automated systems for algorithmic composition intended to reproduce human creativity in some way, there is the problem of evaluating their performance: the concept of artistic creativity eludes a formal, unambiguous and effective definition. This makes it difficult to evaluate these systems in a completely rigorous way. Certainly, many frameworks have been proposed for assessing computational creativity[36], but not one can be easily and uniformly applied to computers and humans alike, in a way that does not spark controversy. It may seem simple to measure computational creativity against human standards: we can simply ask people to listen to human and machine compositions, and declare an algorithmic composition system as creative if these people cannot tell apart its compositions from human ones. As Ariza (2009) noted, this kind of "musical Turing Test" has been performed by many different researchers trying to validate their systems, but they are valid if the algorithmic composition system just aspires to imitate, not to be truly creative and create a truly innovative work of art.

There is also the view that systems for algorithmic composition cannot attain true creativity, even in principle. In fact, it has been suggested (Kugel, 1990) that no Turing-equivalent formalism can truly simulate human creativity, i.e., musical creativity is not effectively computable, thus preventing computer systems from completely imitating human composers, even in theory. This argument is not without merits, but it is open to debate, *precisely* because it lacks a rigorous, unambiguous definition of creativity.

## 4.2 About the Methods

Regardless of these (more or less abstract) considerations about true creativity, this survey has presented existing work on algorithmic composition, organized in several categories. As described at the beginning of Section 3, these categories can be grouped in a few classes:

**Symbolic AI (grammars and rule-based systems).** Under this umbrella, we have grouped very different techniques. These techniques can be used both for imitation (be it the style of a specific composer, or more generally a musical style) and automation of composition tasks. They have proved very effective, and they are very popular (at least, by sheer volume of reviewed work), but in most cases, they are very labor-intensive, because they require musical knowledge to be encoded and maintained in the symbolic framework of choice. There has also been a clear trend towards more and more formal systems, gradually moving from *ad hoc* rule systems to constraint satisfaction and other various formalisms.

---

36. For example, Gero's (2000), Pearce and Wiggins's (2001), Ritchie's (2007) and Boden's (2009).

**Machine learning (Markov chains and artificial neural networks).** Because of their nature, machine learning techniques are used primarily for imitation, although both Markov chains (and related statistical methods) and artificial neural networks can be also used to automate composition tasks (such as harmonization). It should be noted that some techniques described here as symbolic AI are also machine learning (like Cope's ATNs, rule learning or case-based reasoning).

**Optimization techniques (evolutionary algorithms).** As in the case of machine learning, optimization techniques (mostly evolutionary algorithms) have been profusely used for imitation, since it is natural to express the objective of the optimization (the fitness function) as the distance to the musical style to be imitated. However, the automation of composition tasks has also been explored, more so than in the case of machine learning techniques.

**Self-similarity and cellular automata.** Strictly speaking, these techniques are not a form of AI. As explained at the beginning of Section 3, they just represent a convenient way to generate novel musical material without resorting to human musical knowledge, but the problem is that musical material generated in this way is very rough; it is most commonly used by human composers as raw material to build upon.

After reviewing the literature, it becomes apparent that there is no silver bullet: except for strict, limited imitation of specific musical styles or real-time improvisation systems that elaborate on input from human musicians, almost all approaches to algorithmic composition seem to be unable to produce content which can be deemed on a par with professional human composers, even without taking into account the problem of creativity as discussed in Section 4.1. Very few examples stand out, and then only in some niche applications, such as the contemporary classical music composed by Iamus (Ball, 2012).

As there is no silver bullet, one obvious way forward is the hybridization of two or more methods. In fact, from our review of the existing work, it seems apparent that many researchers are already following this route, but there are some hybridizations that have rarely been explored. For example, the music material produced by systems based on self-similarity and CA is commonly regarded as a mere source of inspiration for human composers, rather than as a proper way to automate the composition of music, because this music material generally lacks structure. However, it can be used as the first stage in a process of algorithmic composition, to be modified and refined by subsequent stages, probably based on some form of machine learning if the goal is to produce music in a specific musical style, or some knowledge-based system (such as Leach and Fitch's XComposer, 1995). In the case of evolutionary algorithms, self-similarity systems may be used to seed the initial population, or to introduce variety to avoid premature convergence, and CA may be used as individuals to be evolved (such as in the work of Lo, 2012, which also features machine learning techniques). More research is required to explore the potential of this kind of approach, combining self-similarity and CA-based systems with other methods.

In the case of optimization techniques, the multi-objective paradigm has rarely been used, at least in comparison with the traditional single-objective approach. Composing music usually requires balancing a set of many different, sometimes conflicting objectives, to configure the various aspects of the music, so multi-objective optimization seems a natural way to tackle this problem. All too often, researchers use a weighted sum of parameters to conflate all these objectives into a single fitness function. Very few researchers use multi-objective optimization, as do Geis and Middendorf (2008), Carpentier and Bresson (2010), De Prisco et al. (2010), and Freitas and Guimarães (2011). While multi-objective optimization is harder (both conceptually and in practice), it represents a natural way of dealing with the complexity of having many different objectives, and it should be explored more by the research community.

Finally, in the specific case of evolutionary algorithms, the issue of encoding the individuals should also be examined. Looking at algorithmic composition as an optimization problem, search spaces for musical compositions tend to be huge and high-dimensional. Direct encodings (such as directly representing music as a sequence of pitches) make it very difficult to explore the search space in an effective way, with problems of scalability (the performance degrades significantly as the size of the problem increases) and solution structure (the solutions generated by the algorithm tend to be unstructured, hard to adapt and fragile). This problem is mitigated by indirect encodings, in which the genotype does not directly represent the phenotype, but rather a "list of instructions" to build it. Many different types of indirect encoding have been used in algorithmic composition, such as L-systems, other types of grammars, or the various encoding styles used in genetic programming. However, other advanced techniques for indirect encoding have rarely been applied to algorithmic composition, in order to overcome the aforementioned problems of scalability and solution structure, such as those related to artificial embryogenies (Stanley & Miikkulainen, 2003), which have been inspired by biological developmental processes. Adding these to the evolutionary toolkit may be a way to enable more and more complex compositional tasks to be tackled.

### 4.3 Final Thoughts

Computers have come to stay: the use of CAAC software is prevalent among many composers, and some artistic scenes (as *generative music*) embrace computer-generated music as part of their identity. However, *creativity* is still in the hands of composers for the most part. As argued in Section 4.1, creativity is an inherently subjective concept, and it is arguably debatable the point at which a computational system may become truly creative. However, even if a precise definition cannot be agreed upon, it is easy to see that the development of algorithmic composition systems capable of independent creativity will radically change the process of music composition, and consequently the market for music. This should not be seen as yet another case of computers replacing humans in an ever more sophisticated activity, but a potentially radical disruption in the way composers perform their work: just like a pedagogical expert system does not supersedes the role of human teachers, but enable new ways to do their work.

Being music one of the arts with a stronger mathematical background, it is not surprising that most of the debate on whether machines can make original and creative works has centered in this subfield of computational creativity. Hybridization of different techniques, bioinspiration, and the use of high performance computing might bring about new realms of (computer-) creativity. As science writer Philip Ball put it in his analysis of Melomics' music composition technology: "...unfolding complex structure from a mutable core has enabled the kind of dramatic invention found in biological evolution" (Ball, 2012).

## Acknowledgments

The authors wish to thank Ilias Bergstrom for his comments on a preliminary version of the manuscript. Also, the critical review of our anonymous referees has greatly improved the final version. This study was partially supported by a grant for the MELOMICS project (IPT-300000-2010-010) from the Spanish Ministerio de Ciencia e Innovación, and a grant for the CAUCE project (TSI-090302-2011-8) from the Spanish Ministerio de Industria, Turismo y Comercio. The first author was supported by a grant for the GENEX project (P09-TIC-5123) from the Consejería de Innovación y Ciencia de Andalucía. The first author also wishes to thank his wife Elisa and his daughter Isabel for being there day after day, in spite of the long hours spent writing this manuscript, and his family for the invaluable support they have provided.